\documentclass[review]{elsarticle}
\usepackage[utf8]{inputenc}
\usepackage{mathtools}
\usepackage[english]{babel}
\usepackage{amsthm}
\usepackage{amsmath}
\usepackage{amssymb}
\usepackage{comment}
\usepackage{url}
\usepackage{bbm}
\usepackage{bm}
\usepackage{subcaption}
\usepackage{multirow}
\usepackage{relsize}
\usepackage{color}
\usepackage{wrapfig}
\usepackage[export]{adjustbox}

\newcounter{stepno}

\newtheorem{defn}{Definition}[section]

\newtheorem{lemma}{Lemma}[section]

\newcommand{\E}{\mathbb{E}}
\newcommand{\bn}{\begin{eqnarray}}
\newcommand{\en}{\end{eqnarray}}
\newcommand{\bns}{\begin{eqnarray*}}
\newcommand{\ens}{\end{eqnarray*}}

\newcommand{\defvarbegin}{\begin{quotation}\vspace{-15pt}\begin{tabbing}}
\newcommand{\defvarend}  {\end{tabbing}\vspace{-10pt}\end{quotation}}

\newcommand{\bnarray}{\begin{equation}\begin{array}{rcll}}
\newcommand{\enarray}{\end{array}\end{equation}}

\newcommand{\textwrap}{\parbox[t]{5.5in}}

\newcommand{\barr}{\begin{array}}
\newcommand{\earr}{\end{array}}

\newcounter{cnum}

\newcommand{\beginalg}{\setcounter{stepno}{1}
                \begin{list}{\bf Step~\arabic{stepno}}
                         {\usecounter{stepno}\settowidth{\labelwidth}{\bf Step~9m}
                \addtolength{\leftmargin}{2\parindent}}
                }
\newcommand{\eg}{\end{list}}

\def \define{\begin{quote}\begin{itemize}}
\def \enddefine{\end{itemize}\end{quote}}

\newlength{\boxedparwidth} 
  {\begin{center} \begin{tabular}{|@{\hspace{.15in}}c@{\hspace{.15in}}|}
                 \hline \\ \begin{minipage}[t]{\boxedparwidth}
                 \setlength{\parindent}{0.0in}}%
   {\end{minipage} \\ \\ \hline \end{tabular} \end{center}}
\newcounter{example}
\newcommand{\argmax}{{\rm arg}\max}

\def \xtilde{{\tilde x}}

\def \Ctilde{{\tilde C}}

\def \Itilde{{\tilde I}}

\def \Rtilde{{\tilde R}}
\def \Stilde{{\tilde S}}

\def \phat{\hat p}

\def \Ihat{\hat I}

\def \Mhat{\hat M}

\def \Pbb{\mathbb P}

\def \betahat{\hat \beta}

\def \hbar{\bar h}

\def \pbar{\bar p}

\def \betabar{\bar \beta}

\def \Cbar{\bar C}

\def \Ibar{\bar I}

\def \Rbar{\bar R}
\def \Sbar{\bar S}

\def \Scal{{\cal S}}

\def \Wcal{{\cal W}}
\def \Xcal{{\cal X}}

\def \Zcal{{\cal Z}}

\def \Sbf{{\mathbf S}}
\def \Ibf{{\mathbf I}}
\def \Rbf{{\mathbf R}}

\def \bbf{{\mathbf b}}
\def \Nbf{{\mathbf N}}
\def \xbf{{\mathbf x}}
\def \wbf{{\mathbf w}}
\def \ubf{{\mathbf u}}
\def \vbf{{\mathbf v}}
\def \zbf{{\mathbf z}}
\def \rbf{{\mathbf r}}

\usepackage{lineno,hyperref}
\modulolinenumbers[5]

\journal{European Journal of Operational Research}

\biboptions{authoryear}

\graphicspath{ {./images/} }
\begin{document}

\begin{frontmatter}

\title{Stochastic Optimization for Vaccine and Testing Kit Allocation for the COVID-19 Pandemic}



\author[EEPU]{Lawrence Thul\corref{mycorrespondingauthor}}
\cortext[mycorrespondingauthor]{Corresponding author}
\ead{lathul@princeton.edu}
\tnotetext[t1]{This work is supported by the Air Force Office of Scientific Research (Award Number FA9550-19-1-0203).}
\author[ORFE]{Warren Powell}

\address[EEPU]{Department of Electrical Engineering, Princeton University, Princeton, New Jersey,USA}
\address[ORFE]{Department of Operations Research and Financial Engineering, Princeton University, Princeton, New Jersey,USA}

\begin{abstract}
The pandemic caused by the SARS-CoV-2 virus has exposed many flaws in the decision-making strategies used to distribute resources to combat global health crises.  In this paper, we leverage reinforcement learning and optimization to improve upon the allocation strategies for various resources.  In particular, we consider a problem where a central controller must decide where to send testing kits to learn about the uncertain states of the world (active learning); then, use the new information to construct beliefs about the states and decide where to allocate resources. We propose a general model coupled with a tunable lookahead policy for making vaccine allocation decisions without perfect knowledge about the state of the world.  The lookahead policy is compared to a population-based myopic policy which is more likely to be similar to the present strategies in practice.  Each vaccine allocation policy works in conjunction with a testing kit allocation policy to perform active learning.  Our simulation results demonstrate that an optimization-based lookahead decision making strategy will outperform the presented myopic policy.
\end{abstract}


\end{frontmatter}

\section{Introduction}
During the early months of 2020, it became evident that the SARS-CoV-2 virus was spreading through the global population at an alarming rate.  As the pandemic devastated global economies and supply chains, it was obvious that the strategies in place were not sufficient to handle a crisis at this scale.  Some of the obvious challenges presented by this crisis were the distribution of personal protective equipment, testing kits, and vaccine distribution under extreme shortages, as well as testing accuracy.  At the time of this writing, the world is waiting for a vaccine to suppress the virus in the population and distributing it will present a new series of logistical issues.  This problem is extremely rich with potential to improve distribution processes using quantitative strategies; this work will choose a subset of these problems and utilize reinforcement learning and approximate dynamic programming strategies to improve them.

In this paper, it is assumed that a country/region is in the midst of a pandemic and a decision-maker is tasked with allocating a limited supply of vaccines and testing kits to various local zones in order to try to mitigate the spread of the disease.  These zones could be states, counties, or some other partition of the region.  A key challenge in this setting is the uncertain number of infected individuals, which can only be learned by making observations through testing.  The data collected through testing would then be used to create a probabilistic belief about the true underlying state, which gives vaccine distributors useful information about where to send vaccines to slow the spread.  The combination of making resource allocation decisions under uncertainty, while also performing information collection decisions to improve through active learning, represents a unique modeling and algorithmic challenge.

The heterogeneous characteristics of population densities, health characteristics and behavior will influence transmission rates and virus mobility within local regions; hence, intelligently allocated vaccines will impact the number of infections.  The testing decisions are critical to making the best allocation decisions because the controller needs to have knowledge about the state of the pandemic.  The observations made through testing will reflect biases in the samples caused by factors such as asymptomatic carriers, refusal to get tested, and false positives/negative errors.  It is important to consider the risks caused by the biases because skewed testing data could eventually lead to poor allocation decisions.  The best decisions can be made when each of these characteristics are properly modeled.

This paper makes the following contributions:
\begin{itemize}
    \item It provides the first formal model for a high dimensional pandemic resource allocation problem that captures passive information processes, and active learning..  This model is able to capture the uncertainty in the knowledge about the elements of the SIR model through the belief state, make vaccine allocation decisions under uncertainty, and perform active learning to improve the quality of the belief state.
    \item It proposes a vaccine allocation policy which combines a direct lookahead model with a parameterized approximation of the cost function in order to adjust for risk.  The tunable lookahead policy in conjunction with an active learning policy is capable of outperforming a population-based myopic policy.  We demonstrate this through multiple pandemics simulations on toy networks and simulation models of states in the USA.
    \item We propose an observation function model in order to capture the biases introduced from asymptomatic carriers, refusal to get tested, and false positives/negatives.  The risk-adjusted stochastic lookahead implementation policy proposed in section \ref{section:DLA} can be tuned to reduce the risks posed by biased sampling.
     \item We present a robust modeling framework which is capable of working with environment models of increasing complexity.  As the epidemiological models used to create a simulator improve, the modeling framework has the flexibility to adapt to the new data.
\end{itemize}

The paper is organized as follows.  Section 2 summarizes the literature about vaccine distribution, POMDP modeling, and the unified framework.  Section 3 describes the mathematical model for the environment agent using the unified framework.  Section 4 describes the mathematical model for the controlling agent using the five components of the unified framework.  Section 5 describes implementation policies selected from the two general strategies for designing policies: policy search and lookahead approximations.  Section 5 describes the active learning policy chosen for allocating testing kits.  Section 6 describes the results of the model implemented on an epidemic toy problem and the simulated SARS-CoV-2 pandemic model in the United States.  Section 7 summarizes conclusions from the research.
    
\section{Literature Review}
In section \ref{section:VaccineDist}, we give a brief overview of the literature about vaccine distribution strategies during an epidemic.  In section \ref{section:POMDP}, we give a brief review of partially observable Markov decision processes (POMDP's) and their limitations.  In section \ref{section:unified}, we give an overview about modeling frameworks and their relationship to the unified framework in this paper.

\subsection{Epidemic Planning and Resource Allocation}
\label{section:VaccineDist}

Planning during an epidemic requires making decisions in many different areas such as enacting social distancing measures, contact tracing, vaccine distribution, personal protective equipment distribution, and the allocation of various other essential resources.  Making these decisions can be modeled as sequential decision problems with the common goal of limiting the number of infected individuals.  The literature is rich with policies and algorithms across various different fields to make these decisions intelligently. 

The resource distribution problem is multi-faceted.  The supply chain to deliver resources from the manufacturing facilities to the recipient has many different steps.  For example, some of the decisions which need to be made are where to send resources, who recieves the resources (i.e. who gets vaccinated at a clinic), and how to transport the resources to these locations.  

The implementation decisions in this paper are concerned with where to send vaccines to have the greatest reduction in cases.  Any sequential decision problem requires a dynamic model of the problem and a policy for making the decisions using the model.  One of the most common ways to model a pandemic is to use compartmental models, and these are especially useful because they can be formulated as Markov chains.  In 1927, \cite{SIR_og} created the SIR model which is the most basic compartmental model consisting of three groups within a population: those susceptible (S) to the disease, those infected (I) with the disease, and those removed (R) from the population (from death, recovery, or immunity).  
Since then, the compartmental model literature has expanded past the SIR model to include many different extensions.  Some extensions of the SIR model include: the SEIR (Susceptible-Exposed-Infected-Removed) model, SIRS (Suscepible-Infected-Removed-Susceptible) model, stochastic SIR model, and a spatial SIR model.  The SEIR model is used for diseases with an incubation period before an individual becomes infected.  The SIRS model is used for diseases that do not have permanent immunity.  The stochastic SIR model is used for diseases which have stochastic transmission rates over time.  The spatial SIR formulation is used to model a disease across spatially distributed subpopulations which may interact with one another.  In this paper, we focus on the SIR model with stochastic and spatial extensions; however, there could always be additional extensions which could make it marginally more realistic.  

Most models for planning during epidemics include compartmental models in their logic.  \cite{brandeau_optim} utilizes an SI compartmental model in conjunction with optimization techniques to mitigate the spread of an infectious disease by allocating resources which suppress transmission rates. \cite{Martin1993} uses a spatial SEIR model in conjunction with multiple vaccine allocation strategies to mitigate a measles outbreak amongst dorms on a college campus.  \cite{OptCon_spatial}, \cite{Asano2007}), \cite{MILLER1}, and \cite{SIR_control}  use deterministic spatial SIR models and extensions to derive policies from optimal control theory using variations of the Hamilton-Jacobi-Bellman equations to decide where to allocate vaccines.  \cite{GraphUncertain} formulates a graphical model of a pandemic with uncertainty in the transmission rates reflected through the edges in the graph.  They propose and compare algorithms to allocate a limited number of vaccines across the network to minimize the resulting footprint.  

Another decision is who receives a set of allocated resources.  \cite{geneticAge} uses a stochastic influenza model along with a genetic algorithm to decide which age groups should receive a limited supply of vaccines.  \cite{Medlock1705} uses models of past flu pandemics to compare policies for administering a limited supply of vaccines among various age groups.  \cite{pinar_pandemic} uses a complex spatial SEIR model in conjunction with an age-based component to decide who to feed during a flu pandemic in Georgia by setting up food distribution centers.  

There are various other decisions studied among the literature to optimize a vaccine supply chain.  For example, \cite{coldchain} models a problem to decide whether a distributor will transport vaccines through a cold chain or a non-cold chain to ensure that they are still viable at administration.  \cite{FluOnTime} models an influenza supply chain in the U.S. consisting of healthcare providers, manufacturers, and distributors to ensure on-time delivery to each provider.  

\subsection{The POMDP model}
\label{section:POMDP}
This paper will assume that the true pandemic has a compartmental model; however, we assume that the controller does not have access to this model and must make decisions about where to test to learn about the true state of the pandemic (active learning). If the true state is only partially observable, then the controller must model the uncertainty to make decisions.  There are various modeling strategies to handle decisions under uncertainty.  The next section discusses the modeling framework for partially observable Markov decision processes (POMDP), and its limitations in this problem.

A partially-observable Markov Decision Process (POMDP) is a system which includes a controlling agent making decisions in an environment where the underlying state transitions are Markov; however, the true underlying state of the environment can only be observed partially by the controlling agent.  The finite horizon POMDP problem is modeled using the tuple $(S, \Xcal, \Pbb, R, \Wcal, \Pbb^{obs}, T)$.
\begin{itemize}
\item $S$ - the state space of the environment (e.g. number of infected people in the population).
\item $\Xcal$ - the decision space (e.g. number of vaccines sent to each zone).
\item $\Pbb(s'|s,x)$ - the probability transition matrix for the next environment state given a state-action pair.
\item $R(s,x)$ - the reward function for a given state-action pair.
\item $\Wcal$ - the observation space (e.g. number of new infections measured through testing).
\item $\Pbb^{obs}(w|s,x)$ - the probability of observing $w\in \Wcal$ given a state action pair.
\item T - the time horizon.
\end{itemize}
For the POMDP model, let $s_t$ be the state variable for the environment agent.  The goal of a POMDP is the same goal as an MDP: to find a policy, $X^{\pi}$, which maximizes (or minimizes) the expected sum of cumulative rewards,
\bn
\max_\pi \E^\pi\left\{\sum_{t=0}^T R(s_{t},X^{\pi}_t(s_{t}))|s_0\right\}.
\en
The expectation over $\pi$ suggests that the stochastic processes in the problem are impacted by the decisions made by the policy.  In a POMDP, the true state variable, $s_t$, is unknown to the controlling agent at time $t$, hence the value of the state cannot be computed. If the true state variable is unknown then the state transitions are non-Markovian (from the controlling agents perspective); hence, the entire history of observations and actions would be required to make decisions.  If the entire history of observations had to be stored this would grow extremely quickly.  The rapid growth of the history of actions and observations is known as the curse of history.  

To circumvent the curse of history, the controlling agent stores a belief state variable,  $b_t(s)$  $\forall s\in S$.  $b_t(s)$ is a probability distribution over the (presumably discrete) state space, hence $\sum_{s\in S}b_t(s) =1$.  \cite{belief_proof} shows that the belief state is a sufficient statistic for the history and describes a distribution over the possible states of the environment at time $t$ derived from the previous observations, actions, and states.  It is defined by,
\bns
b_t(s) = \mathbb{P}(s_t|W_{t}, s_{t-1}, x_{t-1}, W_{t-1}, ..., s_0, x_{0}) = \mathbb{P}(s_t|W_{t}, b_{t-1}(s)) .
\ens
When new observations arrive the agent must perform a Bayesian update to compute this new belief state using only the new observations and the previous belief state.  Using the terms in the tuple, we can compute it with, 
\begin{gather}
\label{eqn:POMDP_update}
b_{t+1}(s)=\frac{\mathbb{P}^{obs}(W_{t+1} = w|s', x_t)\sum_{s\in S}\mathbb{P}(s'|s, x_t)b_{t}(s)}{\sum_{s'\in S}\mathbb{P}^{obs}(W_{t+1} = w|s', x_t)\sum_{s\in S}\mathbb{P}(s'|s, x_t)b_{t}(s)}.
\end{gather}
In equation (\ref{eqn:POMDP_update}), the state space must be summed over; if it is a vector then $|\Scal|$ grows exponentially.  Additionally, computing the transition matrix $\Pbb(s'|s,x)$ and observation matrix, $\Pbb(w|s',x)$, when the observations and decisions are vectors will also be intractable.  \cite{POWELL_ADP} discusses the reasons for the intractability of equation (\ref{eqn:POMDP_update}) known as the three curses of dimensionality.
In theory, the solution to the belief MDP can then be used to solve for the optimal policy using Bellman's Equation \cite{POMDP2}.  Let $r(b_t,x_t) = \sum_{s\in S} R(s,x_t)b_t(s)$, then Bellman's optimality equation for a POMDP is given by,
\begin{gather}
\label{eqn:beliefMDP_soln}
\pi^* = \argmax_{x_t} \left(r(b_t, x_t) + \sum_{w\in \Wcal}\Pbb(W_{t+1}=w|s',x_t) V_{t+1}^*(b_{t+1}(s'))\right).
\end{gather}
It is important to note that $b_{t+1}(s')$ must be calculated using equation (\ref{eqn:POMDP_update}).  For finite horizon problems the exact solution is only tractable when the state space, action space, and observation space are small.  The belief state is an uncountable continuous set, therefore trying to find a mapping for each element of this set using backward dynamic programming is not possible.  However, it is a well known result from the POMDP literature that the value function for belief states is still piecewise linear and convex (e.g. \cite{POMDP1}), which implies that an optimal solution exists.  The ability to find the exact optimal solution is almost never possible for real world problems, in fact, the finite horizon POMDP is PSPACE-complete (e.g. \cite{POMDP1}).  \cite{POMDP1} gives a set of solutions using a point-based value iteration approach and compares it to other POMDP solvers.  \cite{POMDP2} derives online planning algorithms for the POMDP problem.  \cite{POMDP_LargeO} formulates strategies for solving POMDPs with continuous and multi-dimensional observations spaces.  \cite{POMDP_beliefComp} attempts to circumvent the curse of dimensionality by finding a low dimensional belief space embedded in the high dimensional belief state to project into.  There have been many algorithms developed for solving POMDPs to make the solutions tractable, but the computational complexities suggest that they would only work for problems with small state, action, and observation spaces compared to the size of the problem discussed in this paper.  

The POMDP modeling approach is widely used for problems with unobservable parameters or quantities, but it suffers from severe computational limitations (it probably cannot be applied to a problem in this paper with more than 3 or 4 zones).  Often overlooked, however, are subtle modeling assumptions that would not apply for our COVID setting.  In particular, the policy in equation (\ref{eqn:beliefMDP_soln}) uses the one-step transition matrix $\Pbb(s'|s,x)$ which, aside from being computationally intractable, implicitly assumes that the transition function is known to the controller.  This means that the controller actually knows the dynamics of how the disease is communicated, which is not the case with COVID-19.  

\subsection{The Unified Framework}
\label{section:unified}
The traditional POMDP formulation is not only computationally intractable for all but toy problems, it does not properly model the fact that the people making decisions about testing and vaccine distribution do not even know the dynamics of the problem.  The POMDP model is a more generalized version of the Markov Decision Process (MDP) framework which forms the roots for much of the reinforcement learning literature.  It is not the only way to model a sequential decision problem in the literature.  

\cite{Powell_Lewis_HB} compares and contrasts reinforcement learning in the MDP framework with the modeling framework in the stochastic optimal control literature.  The discussion by \cite{Powell_Lewis_HB} leads to formulation of the unified framework which ties together the various styles of modeling a sequential decision problem into a consolidated framework which can easily be translated into software.  This framework leads to an optimization problem to search over policies, which discriminates it from other modeling frameworks.
The unified framework consists of five core components: the state variable, the decision variable, the exogenous information process, the transition function, and the objective function.  After the five components are modeled, a policy can be chosen from four classes of policies which are listed below:
\begin{itemize}
    \item Policy function approximation (PFA) - analytical functions directly mapping states to actions, e.g. lookup tables, parametric function,
    \item Cost function approximation (CFA) - an optimization problem involving a parameterized approximation of the cost function, e.g. upper confidence bounding,
    \item Value function approximation (VFA) - approximating the value of a state which can then be utilized in Bellman's equation, e.g. forward approximate dynamic programming,
    \item Direct lookahead approximation (DLA) - developing an approximate model of the future to optimize the best decisions through simulation, e.g. Monte Carlo Tree Search.
\end{itemize}
\cite{Powell_Informs2}, \cite{Powell_Informs}, \cite{PowellEJOR}, and \cite{Powell_RLBook} discuss the formulation of problems using the unified framework in detail.  The unified framework has proven to work in various application domains with high dimensional state and action spaces.  \cite{duranteBADP}, \cite{LinaVehicle}, and \cite{Weidong} utilize the unified framework for problems in application spaces such as energy storage, transportation, and dynamic pricing, respectively.  This paper will view the controller and environment as two separate agents and model each of them with the unified framework.  The interactions between the agents are then modeled as appendages to their respective exogenous information processes.  

\section{Environment Agent Model}
\label{section:env_model}
From the perspective of the controlling agent, the environment agent is a black box which can be queried for observations and impacted by decisions.  The real world will almost always be more complex than any simulator, and a controlling agent would have to approximate the real world to the best of its ability.  Therefore, the environment model was designed to be sufficiently complex to include factors the controlling agent cannot anticipate.  The point is to create a robust model that allows the controlling agent to make effective decisions with an imperfect model.  The model presented in section \ref{section:model} would function with any increasingly complex environment model.  The simulator designed in this section is to make the environment model as realistic as possible by including more stochasticity and complexity than the controlling agent model; hence, we recognize that the simulator constructed for the environment model can be more complex.  

Section \ref{section:VaccineDist} discusses the stochastic spatial SIR model which motivated the design of the environment agent model.  For example, as time passes, there are stochastic interactions between the populations of each zone.  Naturally, zones in close proximity or connected by public transportation are more likely to interact.  Additionally, the system will have an observation function which impacts the samples drawn from the testing centers.  Due to factors caused by inherent characteristics of the disease, human behavior, and the properties of the test there could be biases shifting the distribution of the observation function.  The biases could be caused by factors such as the likelihood of showing symptoms with the disease, the likelihood of going to get tested while showing symptoms, and the errors within the testing kits.  A common issue with some diseases is the lack of symptoms in some portion of the positive cases, also known as an asymptomatic carrier.  The presence of symptoms will cause issues with testing because it decreases the likelihood of a person going to get a test.  To make the testing model as realistic as possible, these factors are included in the sampling oracle.  

The errors within the tests will also be a major concern because they are not always accurate.  There are two types of errors that are possible: false positives and false negatives.  Each type of error will have a probability of occurring given that a test was administered.  Due to limited testing capacity, the number of tests available may exceed the number of people that try to get access to tests.  The model will assume that as the number of tests available increases within a zone, then the likelihood of going to get a test will also increase.

In summary, the environment agent model used to construct the simulator in this paper includes many factors such as spatial correlations between infected populations, biases in testing procedures, stochastically evolving transition rates, and vaccine success rates.  Some of these factors would be extremely difficult to evaluate online as the controlling agent makes decisions, but they impact the results of the simulator.  Hence, we claim that the simulator presented in this section is sufficiently complex to measure the performance of the policies designed in section \ref{section:policy} and the quality of the modeling in section \ref{section:model}.

\subsection{Environment State Variables}
The state variables for the environment agent base model include the variables needed to model the system from time $t$ and onward.  The importance of modeling the simulator and the controller as separate entities is to capture the fact that the controller is not always able to observe the state of the environment perfectly.  Hence, the dynamic state variable for the simulator is given by,
\bn
\label{eqn:state_var}
S_t^{env} = (\Sbf_t, \Ibf_t, \Rbf_t, \alpha_t),
\en
where,
\bns
\Sbf_t &=& \textwrap{vector of the number of susceptible individuals in each zone,}\\
\Ibf_t &=& \textwrap{vector of the number of infected individuals in each zone,}\\
\Rbf_t &=& \textwrap{vector of the number of recovered individuals in each zone,}\\
\alpha_{t} &=& \textwrap{mobility rate at time $t$.}
\ens
Additionally, the initial environment state variable consists of all variables which do not change dynamically, hence it is given by $S_0^{env} = (\Nbf, \gamma, \xi, (\betabar_z)_{z\in \Zcal}, (\sigma_z^\beta)_{z\in\Zcal}, \Sbf_0, \Ibf_0,\Rbf_0)$ where,
\bns
\Nbf &=& \textwrap{vector of zone populations,}\\
\gamma &=& \textwrap{recovery rate of disease,}\\
\xi &=& \textwrap{vaccine success rate,}\\
\betabar_z &=& \textwrap{mean transmission rate of zone $z$,}\\
\sigma_z^\beta &=& \textwrap{variance in transmission rate noise,}\\
\Sbf_0 &=& \textwrap{vector of initial susceptibles,}\\
\Ibf_0 &=& \textwrap{vector of initial infected,}\\
\Rbf_0 &=& \textwrap{vector of initial recovered.}
\ens
In this model of the environment agent, we reiterate that this simulator does not include dynamically changing mean transmission rates, recovery rates, or vaccine success rates.  In the real world, these factors are likely changing dynamically as local laws change and better drugs are developed.  

\subsection{Environment Exogenous Information}
The exogenous information for the environment agent model refer to streams of information that occur exogenously, which we model as stochastic processes.  These variables include:
\bn
\label{eqn:exog_info}
W_{t+1}^{env} = (\xbf_t^{vac}, \Mhat_{t+1}, \hat{\varepsilon}^\beta_{t+1}, \hat{\alpha}_{t+1})
\en
where,
\bns
\xbf_t^{vac} &=& \textwrap{vector of number of vaccines allocated to each zone,}\\
\Mhat_{t+1} &=& \textwrap{random matrix of inter-zone mobility,}\\
\hat{\varepsilon}^\beta_{t+1} &=& \textwrap{additive noise for transmission rate evolution,}\\
\hat{\alpha}_{t+1} &=& \textwrap{mobility rate at time $t+1$,}\\
\hat{n}^{vac}_{t+1} &=& \textwrap{number of vaccines available at time $t+1$.}\\
\ens
\subsection{Environment Transition Functions}
The simulator transition function describes how each simulator state variable evolves between time $t$ and $t+1$.  The spatial SIR model can be modified to allow vaccines (which are managed by the controller) to impact the environment agent.  At time $t$, the $\xbf_t^{vac}$ vaccines pass through the environment transition functions and can impact the true state.  The function of the vaccine is to directly move people from the susceptible group into the recovered/immune group.  The dynamics are captured through the following transition equations,
\bn
\label{eqn:postdecision_trans}
\Sbf^x_{t} &=& (\Sbf_t - \xi\xbf^{vac}_t)^+ \\
\label{eqn:susc_trans}
\Sbf_{t+1} &=& \Sbf^x_{t} - \rbf_{t} \odot \Sbf^x_{t} -\alpha_t \Sbf^x_{t} \odot \Mhat_{t+1}\rbf_{t} \\
\label{eqn:inf_trans}
\Ibf_{t+1} &=& (1-\gamma) \Ibf_{t} + \rbf_{t} \odot \Sbf^x_{t} + \alpha_t \Sbf^x_{t} \odot \Mhat_{t+1}\rbf_{t} \\
\label{eqn:rec_trans}
\Rbf_{t+1} &=& \Rbf_{t} + \gamma \Ibf_t + \xi\xbf^{vac}_t
\en
where $\xi$ is the success rate of the vaccine, $\rbf_{t} = \left[\frac{\betahat_{t+1,z} I_{tz}}{N_z}\right]$, and $\Mhat_{t+1}$ is the mobility probability matrix.    
The mobility rate and number of vaccines available are assumed to be totally exogenous based on the current laws and vaccine manufacturing process, respectively.  The equation for the transition rate at $t+1$ is given by,
\bn
\label{eqn:beta_trans}
\betahat_{t+1,z} = (\betabar_z + \hat{\varepsilon}^\beta_{t+1})^{+}
\en
where, $\hat{\varepsilon}^\beta_{t+1} \sim N(0, \sigma_z^\beta)$.

Since there are no decisions made within the environment there is no objective function. 

\subsection{Environment Observation Function}
\label{section:testing_model}
Another important component of the simulator is to model the impact of testing within each zone.  To make the testing procedure as realistic as possible, the sampling oracle was designed to contain implicit biases. The following definitions are used to outline the important parts of the observation function. 

\begin{defn}
\label{def: Prob}
Assume an individual is randomly drawn from the population of a given zone $z$ at time $t$.  This individual has characteristics related to the possible disease present in the population.  $\Omega^{test} = \{E^{pos}, E^{symp}, E^{test}, E^{tpos}\}$ represents elements of the set of possible outcomes for this individual.  Each element of $\Omega^{test}$ represents the following outcomes:
\bns
E^{pos} &=& \textwrap{\{The person is positive with the disease\},} \\
E^{symp} &=& \textwrap{\{The person is showing symptoms of the disease\},} \\
E^{test} &=& \textwrap{\{The person goes to get tested for the disease between $t$ and $t+1$\},}\\
E^{tpos} &=& \textwrap{\{The person tests positive for the disease\}.}\\
\ens
The complement of each of the outcomes listed are $E^{neg}, E^{asymp}, E^{not}, E^{tneg}$, respectively.
\end{defn}
\begin{defn}
The probability that a person drawn from the population is truly positive with the disease at time $t$ is given by
\bn
p^{inf}_{tz} = \Pbb[E^{pos}]=\frac{I_{tz}}{N_z}.
\en
\end{defn}
When the controlling agent attempts to query a sample from a zone, then the sampled positive cases may not reflect the true percentage in the population, on average.  The controlling agent may not be able to see this bias; however, the environment agent will have access to the information causing the skew in the data. 
\begin{defn}
\label{def:Symp}
Assume a person is drawn from the population and the probability that they are showing symptoms of the disease while carrying it is given by a constant probability $a$, and the probability that they are showing symptoms of the disease while they are not carriers is given by the constant probability $b$.  Hence, $\Pbb[E^{symp}|E^{pos}] = a$, and $\Pbb[E^{symp}|E^{neg}] = b$.
\end{defn}
\begin{defn}
\label{def:cd}
If a person is not showing symptoms while positive with the disease, then they have a lower chance of going to wait in the test queue between time $t$ and $t+1$.  Assume that as the testing capacity of a zone is increased, then the likelihood of getting tested increases through the following equations,
\bn
\label{eqn:c}
c(x_{tz}^{test})  = \Pbb[E^{test} | E^{symp}, x_{tz}^{test}] &=& c_0 + \frac{x_{tz}^{test}}{N_z} (1-c_0), \\
\label{eqn:d}
d(x_{tz}^{test}) = \Pbb[E^{test} | E^{asymp}, x_{tz}^{test}] &=& d_0 + \frac{x_{tz}^{test}}{N_z} (1-d_0),
\en
where $c_0$ and $d_0$ are the initial probability of attempting to get a test while there are none allocated to zone $z$.   
\end{defn}
The simulator is able to access each probability, but the controller would not be able to observe this information directly. It is easier to think about the probabilities $a,b,c$ and $d$ as follows:  $a$ is the probability of being symptomatic while carrying the disease, $b$ is the probability of being symptomatic without carrying the disease, $c$ is the probability of trying to get tested while showing symptoms, and $d$ is the probability of trying to get tested without showing symptoms.  The probabilities $c$ and $d$ are functions of how many tests are available, while $c_0$ and $d_0$ are the base probabilities of trying to get tested given that there are no tests available.  From the controllers perspective, these values cannot be seen however they are reflected in the sample created after administering the test.  

\begin{lemma}
\label{lem:bias_lemma}
Let $f_p$ and $f_n$ be the probability of false positive and false negative, respectively.  $a,b,c(x_{tz}^{test}),d(x_{tz}^{test})$ are from definitions \ref{def:Symp} and \ref{def:cd}.  $p_{tz}^{test}$ represents the probability of testing positive if the person gets tested given by,
\bns
p_{tz}^{test} &=& (1-f_n) \Pbb[E^{pos}|E^{test}] + f_p (1-\Pbb[E^{pos}|E^{test}]),
\ens
where,
\bns
\Pbb[E^{pos}|E^{test}, x_{tz}^{test}] =\frac{\left(ac + (1-a)d\right)p^{inf}_{tz}}{(c - d)\left((a-b)p^{inf}_{tz} + b\right) + d}.
\ens
The variables $c$ and $d$ are functions of the number of tests allocated to a zone. \begin{proof}
See Appendix.
\end{proof}
\end{lemma}

The adjusted probability allows the factors affecting the observation distribution to be reflected in the probability of testing positive.  Therefore, the number of positive test results after administering the limited number of tests, $x_{tz}^{test}$, is given by the random variable, $\Ihat_{t+1,z}$.  The random variable $\Ihat_{t+1,z}$ has a binomial distribution with parameters $(n,p) = (x_{tz}^{test},p_{tz}^{test})$.  Lemma \ref{lem:bias_lemma} implies the parameter of the binomial distribution for the observation function does not reflect the true percentage of infected individuals in the population on average, but a shifted distribution with a bias unknown to the controller.

\section{Controlling Agent Model}
\label{section:model}
This section proposes a model for the controlling agent using the five components of the unified framework.  The controller does not have access to the environment agent's state variable, $S_t^{env}$, or the transition functions describing how they evolve, $S^{M,env}(S_{t}^{env}, W^{env}_{t+1})$.  Due to the lack of this necessary information, the controller must make assumptions about the components of the environment agent's model.  We are going to start by outlining the belief model, before giving the five elements of the controller model.

\subsection{Belief Model}
\label{section:belief}
In a sequential decision problem with imperfectly known states and state transitions, the controller must maintain a belief state and have a belief about the transition functions in the environment mode, which are likely unknown.  This section contains the three major components of the belief model:
\begin{itemize}
    \item[1)] the belief state,
    \item[2)] the belief about the transition functions in the environment model,
    \item[3)] the updating equations for the belief state.
\end{itemize}

\subsubsection{Belief State}
The belief state contains the parameters of a probability distribution over the environment agent's state space.  In the SIR model, there are three physical state variables describing the population of susceptible (S), infected (I), and removed individuals (R).  The S,I, and R subpopulations are elements of the environment agent state, and are not known perfectly to the controlling agent.  Since the controlling agent cannot observe the environment state perfectly at time $t$ it must maintain a probability distribution over the entire state space.  The controlling agent can iteratively update the belief state as it queries observations from the testing centers.

\begin{defn}
The SIR model assumes that the total population remains constant.  Therefore, the belief about the true percentage of each subpopulation in a zone has the following property,
\bns
\label{eqn:constraint_pop}
\pbar^{susc}_{tz} + \pbar^{inf}_{tz} + \pbar^{rec}_{tz} = 1.
\ens
The property implies that the belief about the subpopulation of each zone has a multinomial distribution with parameters $(S_{tz}, I_{tz}, R_{tz}) \sim Mult(N_z, \pbar^{susc}_{tz}, \pbar^{inf}_{tz}, \pbar^{rec}_{tz})$.  Therefore, the belief state will contain the dynamically changing parameters of the multinomial distribution for each zone.
\end{defn}

\subsubsection{Environment Transition Function Belief}
\label{section:Etrans_belief}
The controlling agent assumes a functional form for the transition functions in the environment model.  The transition functions in the environment model are used to update the state variable known perfectly to the environment agent.  Therefore, the functions in this section would describe the set of equations to update the state variables if there was no uncertainty in the state at time $t$.  

Let $(\Sbf_{t}, \Ibf_{t}, \Rbf_{t})$ be vectors containing the number of susceptible, infected and removed individuals in each zone $z$, respectively.  The vectors are the physical state variables for the environment agent.  If the controlling agent had access to the true physical state variables and made the decision $\xbf_{t}^{vac}$, then it believes the environment will transition to the next state according to the following set of equations:
\bn
\label{eqn:bstate_trans_s}
\Sbf_{t+1} &=& f^{susc}(\Sbf_{t}, \Ibf_{t}, \xbf_{t}^{vac}|S_{0}^{env}), \\
\label{eqn:bstate_trans_i}
\Ibf_{t+1} &=& f^{inf}(\Sbf_{t}, \Ibf_{t}, \xbf_{t}^{vac}|S_{0}^{env}), \\
\label{eqn:bstate_trans_r}
\Rbf_{t+1} &=& f^{rec}(\Rbf_{t}, \Ibf_{t}, \xbf_{t}^{vac}|S_{0}^{env}).
\en
Equations (\ref{eqn:bstate_trans_s} - \ref{eqn:bstate_trans_r}) describe the belief about how each environment state variable evolves.  Since the true state variable is not fully observable each element of $(\Sbf_{t}, \Ibf_{t}, \Rbf_{t})$ are random variables; hence, the function cannot be evaluated by the controlling agent because it only has a belief state describing the random variables.  However, the equations are important because they must be used to derive the updating equations for the belief state variables.  

\subsubsection{Belief State Update}
The controlling agent will decide to allocate $x_{tz}^{test}$ testing kits to zone $z$ at time $t$.  Between time $t$ and $t+1$ the testing kits are administered and produce a random response, $\Ihat_{t+1,z}$, which represents the total number of positive cases measured in zone $z$ between $t$ and $t+1$.  Assume the random variable $\Ihat_{t+1,z}$ has a binomial distribution with parameters $(n,p) = (x_{tz}^{test}, p_{tz}^{test})$.  The response would be analogous to the output of the observation function defined in the POMDP model and the specifics are outlined in section \ref{section:testing_model}.  

The probability, $p_{tz}^{test}$, is the probability of randomly selecting an infected individual out of the population of individuals who received a test at time $t$ in zone $z$.  This probability is affected by factors such as the likelihood of going to get a test while showing symptoms, the likelihood of showing symptoms while positive and the probability of false positives and false negatives.  The exact impact those factors will have on $p_{tz}^{test}$ is not known to the controller, but the policies we discuss later will try to make decisions which can mitigate the risks imposed by the sampling biases.

The belief state updating equations take the observations, $\Ihat_{t+1,z}$, queried from the testing centers and use them to estimate the new belief state at $t+1$.  Each of the three parameters $\pbar_{t+1,z}^{susc}$, $\pbar_{t+1,z}^{inf},$ and $\pbar_{t+1,z}^{rec}$ in each zone $z$ will need an updating equation.  Note that the infected population is being observed directly with $\Ihat_{t+1}$, but the susceptible and removed populations are not.

Let $\alpha_{tz}^{prior}$ and $\beta_{tz}^{prior}$ be parameters of a beta distribution which is constructed to encode the prior information about the number of infected individuals in the population of zone $z$.  The controlling agent has prior knowledge about the infected population prior to the observations at $t+1$ because of the assumed functions in section $\ref{section:Etrans_belief}$.  The prior information can be encoded at $t+1$ by taking the expectation of equation $\ref{eqn:bstate_trans_i}$ given by,
\bn
\label{eqn:beta_prior_a}
\alpha^{prior}_{tz} &=& \E[I_{t+1,z}|S^{cont}_{t}], \\
\label{eqn:beta_prior_b}
\beta^{prior}_{tz} &=& N_z - \E[I_{t+1,z}|S^{cont}_{t}].
\en
The observations are assumed to be drawn from a binomial distribution with parameters $(x_{tz}^{test}, p_{tz}^{test})$.  $p_{tz}^{test}$ is unknown to the controller, but a sample was just observed from its distribution $\Ihat_{t+1,z}$.  The beta prior is a conjugate prior to the binomial distribution, and through Bayes' theorem the updated distribution will be beta-binomial.  The resulting distribution can be thought of as the binomial parameter $p_{tz}^{test}$ being drawn from the beta distribution with parameters  $\alpha_{tz}^{prior}$ and $\beta_{tz}^{prior}$. 
\begin{lemma}
\label{lem:beta_binomial}
Let $\alpha_{tz}^{prior}$ and $\beta_{tz}^{prior}$ be parameters of a beta distribution and $\Ihat_{t+1,z}$ be a sample drawn from a binomial distribution with $x_{tz}^{test}$ trials.  The compound distribution produced by Bayes' Theorem is a beta-binomial distribution.  The estimator for the probability of an infection, $\pbar_{t+1,z}^{inf}$ is given by,
\bn
\label{eqn: beta_bin}
\pbar^{inf}_{t+1,z} = \frac{ \Ihat_{t+1,z} + \alpha^{prior}_z}{x_{tz}^{test} +\alpha^{prior}_z +\beta^{prior}_z}.
\en
\begin{proof}
See Appendix.
\end{proof}
\end{lemma}
After the tests have been administered into the population, it is possible to get an estimate of the number of infected individuals; however, there are two other groups in the population: susceptible and removed.  Since we only have observations of the number of infected individuals at time $t+1$, then it would not be possible to estimate the other two groups without using the belief about the transition functions in section \ref{section:Etrans_belief}.  To estimate the other two subpopulations within a zone, the belief about the environment transition functions must be used to update the belief states for susceptible and removed individuals.  

After the observations have streamed into the system from the testing centers, the parameters of the beta distribution are computed using the belief about the transition equations and used to get the posterior parameter of the beta-binomial distribution given in equation (\ref{eqn: beta_bin}).  The posterior distribution for the infected number of individuals is used to update the parameters of the multinomial distributions representing the susceptible and removed subpopulations within each zone.  
\begin{defn} 
\label{def:proj}
Let $\Pi_{\Delta_z}$ be the projection operator for the set defined by,
\bns
\Delta_{z} = \left\{(\pbar_{t+1,z}^{susc}, \pbar_{t+1,z}^{rec}) \in [0,1]^2: \pbar_{t+1,z}^{susc} +  \pbar_{t+1,z}^{rec} = 1 - \pbar_{t+1,z}^{inf}\right\}.
\ens
\end{defn}
The expected susceptible and removed subpopulations of each zone is computed by taking the expectation of equation (\ref{eqn:bstate_trans_s}) and (\ref{eqn:bstate_trans_r}), respectively.  If the terms are not in the set $\Delta_z$ in definition \ref{def:proj}, then they must be projected back to the nearest point.  This set of equations is given by,
\bn
\label{eqn:comp_sbar}
\Sbar_{t+1,z} &=& \E[S_{t+1,z}|S_t^{cont}, \xbf_{t}^{vac}], \\
\label{eqn:comp_rbar}
\Rbar_{t+1,z} &=& \E[R_{t+1,z}|S_t^{cont}, \xbf_{t}^{vac}], \\
\label{eqn:proj_SR}
(\pbar_{t+1,z}^{susc}, \pbar_{t+1,z}^{rec}) &=& \Pi_{\Delta_z}\left[\left(\frac{\Sbar_{t+1,z}}{N_z}, \frac{\Rbar_{t+1,z}}{N_z}\right)\right].
\en

In summary, the controlling agent updates the parameters in the belief state through the following process:
\begin{itemize}
    \item[1)] Make observations $\Ihat_{t+1,z}$ for all $z \in \Zcal$,
    \item[2)] Compute the estimator of $\pbar_{t+1,z}^{inf}$ in equation (\ref{eqn: beta_bin}),
    \item[3)] Use equations (\ref{eqn:comp_sbar} - \ref{eqn:proj_SR}) to update $\pbar_{t+1,z}^{susc}$ and $\pbar_{t+1,z}^{rec}$.
\end{itemize}
It is important to emphasize the beliefs $\pbar_{t+1,z}^{susc}$ and $\pbar_{t+1,z}^{rec}$ would not be possible to update through observations of $\Ihat_{t+1}$ without the belief about the transition equations presented in section \ref{section:Etrans_belief}.  The belief about the transition equations may not match the true model (as we demonstrate in this paper) because it is usually an approximation of the real world, which is usually not possible to know perfectly.  After the belief model is fully defined, then the five components of the unified framework naturally follow.  
\subsection{State Variables}
\label{section: state}
The state variables include the information which is needed to compute the transition functions, objective function, and policy for making decisions at time $t$.  Any information which is not changing dynamically remains a latent variable defined in the initial state.  The state variable for the base model is defined as,
\bn
\label{eqn:state}
S^{cont}_{t} = ((\pbar^{inf}_{tz},\pbar^{rec}_{tz},\pbar^{susc}_{tz})_{z\in\Zcal},  n_{t}^{vac}, n_t^{test})
\en
where,
\bns
\pbar^{inf}_{tz} &=& \textwrap{the estimated infected population \% at time $t$ in zone $z$,} \\
\pbar^{rec}_{tz}&=& \textwrap{the estimated recovered population \% at time $t$ in zone $z$,} \\
\pbar^{susc}_{tz}&=& \textwrap{the estimated susceptible population \% at time $t$ in zone $z$,} \\
n_{t}^{vac}&=& \textwrap{the number of vaccines available at time $t$,} \\
n_t^{test}&=& \textwrap{the number of tests available at time $t$.} \\
\ens
Let the state space be denoted by $\Scal^{cont}$.  
The initial state contains the variables which are not changing dynamically.  The initial state for this model is given  by,
\bns
S^{cont}_{0} = ((\beta_z, \gamma_z, \pbar^{inf}_{0z},\pbar^{rec}_{0z},\pbar^{susc}_{0z})_{z\in\Zcal}, n_{0}^{vac}, n_0^{test})
\ens
where,
\bns
\beta_z &=& \textwrap{the transmission rate of each zone $z$,} \\
\gamma_z &=& \textwrap{the recovery rate of each zone $z$,} \\
\pbar^{inf}_{0z} &=& \textwrap{initial belief of the infected \% in zone $z$,} \\
\pbar^{rec}_{0z}&=& \textwrap{initial belief of the recovered\% in zone $z$,} \\
\pbar^{susc}_{0z}&=& \textwrap{initial belief of the susceptible \% in zone $z$,} \\
n_{0}^{vac}&=& \textwrap{the number of vaccines available at time $t=0$,} \\
n_0^{test}&=& \textwrap{the number of tests available at time $t=0$.}
\ens

\subsection{Decision Variables}
There are two sets of decisions at each time step: the information collection decision of how many tests to allocate and the implementation decision of how many vaccinations to allocate to each zone.   The decision to allocate vaccines to each zone is given by, $x_{tz}^{vac}$, which is constrained by the total number of vaccines available, $n_t^{vac}$.  The decision to allocate testing kits to each zone is given by, $x_{tz}^{test}$, which is constrained by the total number of tests available, $n_t^{test}$.  Hence, the vaccine decision variable is constrained by,
\bn
\sum_{z\in\Zcal} x_{tz}^{vac} &\leq& n_t^{vac}.
\en
The testing decision variable is constrained by,
\bn
\sum_{z\in\Zcal} x_{tz}^{test} &\leq& n_t^{test}.
\en

The set of decisions can be characterized by $\xbf_t = (\xbf_{t}^{vac}, \xbf_{t}^{test})$.

\subsection{Exogenous Information}
The exogenous information, $W^{cont}_{t+1}$, represents all the infromation that arrives between time $t$ and $t+1$.  It is given by,
\bn
W_{t+1}^{cont} = ((\Ihat_{t+1,z})_{z\in\Zcal}, n_{t+1}^{vac}, n_{t+1}^{test})
\en
where,
\bns
\Ihat_{t+1,z} &=& \textwrap{number of positive tests sampled,} \\
n_{t+1}^{vac} &=& \textwrap{the number of vaccines available at $t+1$,}\\
n_{t+1}^{test} &=& \textwrap{the number of tests available at $t+1$.}
\ens
The number of tests and vaccines available at $t+1$ is a process evolving completely exogenously from the model.  These values could be determined by the bottlenecks in the manufacturing processes from the distributors.  It is assumed that at the beginning of each time step this number is revealed to the system; however, it does not attempt to model the stochastics behind it.

\subsection{Transition Function}
The transition function is the set of equations which describe how each of the state variables evolve between time $t$ and $t+1$.  The state variable is given in Equation (\ref{eqn:state}).  A simple belief about the transition function would assume the basic SIR model where each zone is evolving independently.  Equations (\ref{eqn:bstate_trans_s} - \ref{eqn:bstate_trans_r}) from section \ref{section:Etrans_belief} are defined as,
\bn
\label{eqn:bstate_trans_sx_base}
\Sbf^x_{t} &=& (\Sbf_t - \xbf_{t}^{vac})^+, \\
\label{eqn:bstate_trans_s_base}
\Sbf_{t+1} &=& \Sbf^x_{t} - \bbf \odot \Sbf_t^x \odot \Ibf_t,\\
\label{eqn:bstate_trans_i_base}
\Ibf_{t+1} &=& (1-\gamma) \Ibf_{t} + \bbf \odot \Sbf_t^x \odot \Ibf_t,\\
\label{eqn:bstate_trans_r_base}
\Rbf_{t+1} &=& \Rbf_{t} + \gamma \Ibf_t + min(\Sbf_t, \xbf_t^{vac}),
\en
where $\bbf = \left[\frac{\beta_z}{N_z}\right]_{z\in\Zcal}$ and $\odot$ is the elementwise vector multiplication operator.  The variables in equations (\ref{eqn:bstate_trans_sx_base} - \ref{eqn:bstate_trans_r_base}) are not known to the controller, hence they must be estimated statistically.  
\begin{lemma}
\label{lem:itrans_lem}  The belief state for each zone is represented by a multinomial distribution with parameters $(N_z,\pbar^{inf}_{tz},\pbar^{susc}_{tz},\pbar^{rec}_{tz})$.  The expectation of equations (\ref{eqn:bstate_trans_sx_base} - \ref{eqn:bstate_trans_r_base}) do not have a closed form.  If the populations of the zones are fairly large $(N_z \gtrapprox 100)$, then these expectations are approximated by assuming each subpopulation of a zone is normally distributed and independent.  We use bars to denote an approximated expectation (e.g. $\Sbar \approx \E[S]$).  Through this assumption, the estimates of the expectation of equations (\ref{eqn:bstate_trans_sx_base} - \ref{eqn:bstate_trans_r_base}) are given by,
\bn
\label{eqn:susxbar_t1}
\Sbar_{tz}^x &=& \left( (\Sbar_{tz} -x_{tz}^{vac})\Phi\left(\frac{\Sbar_{tz} -x_{tz}^{vac}}{\sigma^{susc}_{tz}}\right) + \sigma^{susc}_{tz}\phi\left(\frac{\Sbar_{tz} -x_{tz}^{vac}}{\sigma^{susc}_{tz}}\right)\right), \\
\label{eqn:susbar_t1}
\Sbar_{t+1,z} &=& \left(1 - \beta_z \pbar_{tz}^{inf}\right) \Sbar_{tz}^x, \\
\label{eqn:infbar_t1}
\Ibar_{t+1,z} &=& (1-\gamma)\Ibar_{tz} + \beta_z \pbar_{tz}^{inf} \Sbar_{tz}^x, \\
\label{eqn:recbar_t1}
\Rbar_{t+1,z} &=& \Rbar_{tz} + \gamma \Ibar_{tz} +\Sbar_{tz} - \Sbar_{tz}^x,
\en
where,
\bns
\Sbar_{tz} &=&  N_z \pbar^{susc}_{tz}, \\
\sigma^{susc}_{tz} &=&  \sqrt{N_z \pbar^{susc}_{tz} (1-\pbar^{susc}_{tz})}, \\
\Ibar_{tz} &=&  N_z \pbar^{inf}_{tz}, \\
\Rbar_{tz} &=&  N_z \pbar^{rec}_{tz},
\ens
and $\Phi$ is the standard normal cdf and $\phi$ is the standard normal pdf.
\begin{proof}
See Appendix.
\end{proof}
\end{lemma}

If $\alpha^{prior}_{tz} = \Ibar_{t+1,z}$ and $\beta^{prior}_{tz} = N_z - \Ibar_{t+1,z}$, then Lemma \ref{lem:beta_binomial} gives the updating procedure for $\pbar_{t+1,z}^{inf}$.

The other belief state variables, $\pbar_{t+1,z}^{rec}$ and $\pbar_{t+1,z}^{susc}$, are computed using equations (\ref{eqn:comp_sbar} - \ref{eqn:proj_SR}) and the result of Lemma \ref{lem:itrans_lem}.  Hence,
\bn
(\pbar_{t+1,z}^{susc}, \pbar_{t+1,z}^{rec}) &=& \Pi_{\Delta_z}\left[\left(\frac{\Sbar_{t+1,z}}{N_z}, \frac{\Rbar_{t+1,z}}{N_z}\right)\right].
\en

\subsection{Objective Function}
The objective function measures the performance of the system.  It will measure how well the vaccine and testing allocation decisions impact the state of the epidemic.  The goal of the problem is to come up with a decision making strategy, or policy, to minimize the total number of infected people in the population throughout the entire time horizon.  Therefore, the cumulative reward objective can be posed as the following optimization problem,
\bn
\min_{\pi \in \Pi} \E \left\{\sum_{t=0}^{T-1} \Cbar(S^{cont}_{t},X^{\pi}(S_t^{cont})) | S_0^{cont}\right\}.
\en
The function $\Cbar(S^{cont}_{t},X^{\pi}(S_t^{cont}),W_{t+1})$ is a one period cost function describing the metric which is evaluated by the controller at each time period.  The goal of mitigating the spread of disease is to minimize the number of infected people, but it is not possible to perfectly observe the true number of infected people at $t+1$.  If the controller had access to the true state and transition equations, then the goal would be to minimize the true one-step cost given by,
\bn
\label{eqn:perfect_goal}
C^{env}(S^{env}_{t},\xbf_t,W^{env}_{t+1}) = \sum_{z\in\Zcal} I_{t+1,z}(x_{tz}^{vac}).  
\en
Equation (\ref{eqn:perfect_goal}) is not observable, hence the controller will minimize the expected number of individuals infected with the disease, given the current belief state.  The belief state contains the parameters of the probability distribution over the environment state variables, hence the controllers cost function is given by,
\bn
\label{eqn:cost_fun}
\Cbar(S^{cont}_{t},\xbf_t) = \sum_{z\in\Zcal} \E[I_{t+1,z}(x_{tz}^{vac})|S_{t}^{cont},x_{tz}^{vac}] = \sum_{z\in\Zcal} \Ibar_{t+1,z}(x_{tz}).
\en
To evaluate this cost function, the belief about the transition function for the infected number of individuals must be used to predict what the impact of the decision will be, hence equation (\ref{eqn:infbar_t1}) must be used to evaluate the right hand side of equation (\ref{eqn:cost_fun}).

\section{Designing Policies}
\label{section:policy}
The policy is a mapping from the state space to the decision space.  At time $t$ there are implementation decisions (the number of vaccines to allocate to each zone) and learning decisions (the number of testing kits to allocate to each zone).  In section \ref{section:vaccine_dec}, we illustrate two types of implementation policies: one from the PFA class and one from the DLA class.  In section \ref{section:test_dec}, we present a CFA learning policy for deciding the zones with the most valuable information to collect through testing.

\subsection{Implementation Policies}
\label{section:vaccine_dec}
The implementation decision in this problem setting is to choose how many vaccines to send to each zone in the nation, $x_{tz}^{vac}$.  The decision space for the next set of policies is given by,
\bn
\label{eqn:dec_set}
\Xcal_t^{vac} = \{\xbf_t^{vac} \in\mathbb{N}^{|\Zcal|}: \mathbf{1}^T \xbf_t^{vac} \leq n_{t}^{vac}\}.
\en
The optimal policy for making a decision is given by Bellman's equation defined in equation (\ref{eqn:beliefMDP_soln}), 
\bn
X_t^*(S_t^{cont}) = \arg\min_{\xbf_t^{vac} \in \Xcal_t^{vac}} \left(\Cbar(S^{cont}_{t},\xbf_t) + \E_{W_{t+1}}[V^*_{t+1}(S^{cont}_{t+1})|S^{cont}_{t}]\right)
\en
where $V^*_{t+1}(\cdot)$ is the optimal value function at $t+1$ if we knew the true state of the environment.  The dimension of the state variable defined in section \ref{section: state} is $3|\Zcal| + 2$, therefore as $|\Zcal|$ grows it becomes computationally intractable to solve for the optimal policy due to the dimensionality of the state variable.  When the optimal policy is not possible to achieve, then the goal of the problem becomes finding the best approximation to it.  When designing the approximation, there are four possible classes of policies to choose from, which are defined in section \ref{section:unified}.

\subsubsection{PFA: Population-based allocations}
The PFA class of policies is an analytical function which maps states to actions, and usually includes tunable parameters.  These policies can be used to characterize a simple rule-based method, which is why these are most commonly used in practice.  Some examples of very simple PFA policies could be to allocate the set of vaccines evenly among the zones, proportionally by population, or proportionally by the belief of the number of infected individuals.  In this paper, the PFA we chose to analyze consists of a sigmoid function, which is a function of $\pbar_{tz}^{susc}$, with tunable parameters $\theta^{PFA} = (\theta^{PFA}_0, \theta^{PFA}_1)$.  The sigmoid function for each zone is given by,
\bns
g_z(S_{t}^{cont}|\theta^{PFA}) = \left(\frac{1}{1+e^{-\theta_1^{PFA} \pbar_{tz}^{susc}}}\right)^{\theta_0^{PFA}}.
\ens
Then, to proportionally allocate each set of responses and ensure that the solution remains in the decision set, the policy becomes:
\bn
\label{eqn:PFA}
X^{PFA}(S_{t}^{cont}|\theta^{PFA}) =  \left[\Big\lfloor \frac{n_{t}^{vac}g_z(S_{t}^{cont}|\theta^{PFA})}{\sum_{z\in \Zcal}g_z(S_{t}^{cont}|\theta^{PFA})} \Big\rfloor \right]_{z\in \Zcal}.
\en

This policy operates quickly online, but it requires a time consuming offline grid search to find the best set of hyperparameters.  The policy will proportionally allocate the vaccines based on the magnitude of the response caused by the belief about the percentage of the population that is susceptible to the disease.

\subsubsection{DLA: Parameterized Two-Step Lookahead Approximation}
\label{section:DLA}
In a direct lookahead approximation, an approximate model of the future is created to make the best decisions at $t$ by looking at the impact of decisions in the future.  The lookahead model will approximate the base model by choosing the $\theta^{risk}$-percentile of the distribution represented by the distribution of the belief state, and using this value as the lookahead state variable. The lookahead model will simulate the future by using the belief about the transition functions in equations (\ref{eqn:bstate_trans_sx_base} - \ref{eqn:bstate_trans_r_base}).  This is a reasonable approximation of the base model; however, it omits performing active learning.  To create a proper lookahead model the five components of a sequential decision model defined by the unified framework in section (\ref{section:unified}) must be defined.  A lookahead variable will have a tilde and two time indices, where the first is the time within the base model and the second index is the time within the lookahead model.

\paragraph{Lookahead State Variable}
The distribution of each subpopulation of each county is approximated by a normal distribution defined in Lemma \ref{lem:itrans_lem}; hence the $\theta^{risk}$-percentile of the susceptible distribution is given by $\Stilde^{\theta}_{tt} = \Sbar_{tz} - z^{\theta}\sigma_{tz}^{susc}$.  The number of infected individuals will be approximated by the mean of the distribution $\Itilde_{ttz} = \Ibar_{tz}$, and the number of removed individuals will be $\Rtilde^{\theta}_{ttz} = N_z -\Stilde^{\theta}_{ttz} -\Itilde_{ttz}$.  The model at time $t$ does not have access to a forecast of the number of vaccines available in the future, hence the lookahead model will assume that the number of vaccines available at each $t'$ in the future, where $t'\geq t$, will be the same.  Hence, the lookahead state for $t'$ in the future is given by,
\bns
\Stilde^{cont}_{tt'}(\theta^{risk}) =\left(\Stilde^{\theta}_{tt'z}, \Itilde_{tt'z}, \Rtilde^{\theta}_{tt'z}\right)_{z \in \Zcal}
\ens

\paragraph{Lookahead Decisions}  The lookahead model will not attempt to learn in the future, hence the testing decision will not be modeled.  Therefore, the only lookahead decision is the decision about where to allocate vaccines, $\mathbf{\xtilde}_{tt'}$.  Each of these decision vectors is constrained by $\mathbf{1}^T\mathbf{\xtilde}_{tt'} \leq n_t^{vac}$.

\paragraph{Lookahead Exogenous Information} The lookahead model is using point estimates derived from the belief state, and the belief transition functions for a point estimate are deterministic, hence there is not an exogenous information variable.

\paragraph{Lookahead Transition Functions} The lookahead transition functions are given by Equations (\ref{eqn:bstate_trans_sx_base}-\ref{eqn:bstate_trans_r_base}), which are deterministic functions.  The lookahead transition equations allow the future to be approximated as long as the initial lookahead state, $\Stilde_{tt}^{cont}$, is in the feasible state space. 

\paragraph{Lookahead Objective Function}  The lookahead objective function is given by
\bn
\label{eqn:LA_goal}
\Ctilde(\Stilde^{cont}_{tt'},\mathbf{\xtilde}_{tt'}) = \sum_{z\in\Zcal} \Itilde_{t,t+1,z}(\xtilde_{tt'z}).  
\en

The optimization problem for a two-step lookahead approximation is given by,
\bn
\label{eqn:2step_obj}
X^{DLA,2}_{t} = \arg\min_{\mathbf{\xtilde}_{tt}, \mathbf{\xtilde}_{t,t+1}} \Ctilde(\Stilde_{tt}^{cont},\mathbf{\xtilde}_{tt})  + \Ctilde(\Stilde_{t,t+1}^{cont}, \mathbf{\xtilde}_{t,t+1})
\en

The optimization problem in equation (\ref{eqn:2step_obj}) reduces to a problem with a nonconvex quadratic objective function with linear constraints.  For problems where $|\Zcal|$ is not too large the policy can be solved with bilinear solvers.  The details describing the solution to equation (\ref{eqn:2step_obj}) can be found in the  Appendix.

\subsection{Learning Policies}
\label{section:test_dec}
The second type of decision is to select how many samples to observe from each zone through testing.  At time $t$ the controlling agent must decide which zones to send the $n_t^{test}$ kits after the implementation policy has already been made.  The learning decision will impact the distribution of the random variable for positive tests, $\Ihat_{t+1}$, and consequently affect the future decisions about where to send vaccines.

It is important to have estimates of each environment state variable, and this decision will change those estimates.  The testing policy we propose consists of a tradeoff between proportionally allocating a percentage of the testing kits, and trying to minimize the variance in the belief state at $t+1$.  The logic behind this design is get close to an estimate with respect to mean-squared error globally in the region, but also ensure that each zone is being sampled. 

Equation (\ref{eqn: beta_bin}) updates the belief about $p^{inf}_{t+1}$ by assuming the posterior has a beta-binomial distribution.  If the estimator in equation (\ref{eqn: beta_bin}) has a beta-binomial distribution, then its variance is given by,
\bn
\label{eqn:beta_bin_var}
Var[\pbar_{t+1,z}^{inf}|S^{cont}_t] = \frac{x^{test}_{tz} \alpha_{tz}^{prior} \beta_{tz}^{prior} ( \alpha_{tz}^{prior} + \beta_{tz}^{prior} + x^{test}_{tz}) }{( \alpha_{tz}^{prior} + \beta_{tz}^{prior})^2 ( \alpha_{tz}^{prior} + \beta_{tz}^{prior} + 1)}.
\en
Let the sum of variances in equation (\ref{eqn:beta_bin_var}) become the objective function.  The testing policy is induced by the optimization problem when the sum of variances is the objective function, given by
\bn
\label{eqn:obj_test}
\min_{\xbf^{test}} \sum_{z\in\Zcal} Var[\pbar_{t+1,z}^{inf}(\xbf^{test})|S^{cont}_t],
\en
\bns
\text{subject to} \hspace{3mm} \mathbf{1}^T \xbf^{test} \leq n_t^{test}.
\ens
\begin{lemma}
\label{lem:learning_pol}
Let equation (\ref{eqn:obj_test}) be the optimization problem used to produce the learning decisions.  The optimization problem reduces to a quadratic program with linear constraints given by,
\bn
\label{eqn:test_quad}
\min_{\xbf^{test}} (\xbf^{test})^T D \xbf^{test} + d^T \xbf^{test},
\en
\bns
\text{subject to} \hspace{3mm} \mathbf{1}^T \xbf^{test} \leq n_t^{test}.
\ens
where,
\bns
D &=& diag\left(\left[\frac{\alpha_{tz}^{prior} \beta_{tz}^{prior}}{N_z^2 ( N_z + 1)}\right]_{z\in\Zcal}\right)\\
d &=& \left[\frac{\alpha_{tz}^{prior} \beta_{tz}^{prior}}{N_z ( N_z + 1)}\right]_{z\in\Zcal}.
\ens
This is an integer quadratic program with linear constraints. 
\end{lemma}
The problem with the policy induced by equation (\ref{eqn:test_quad}) is it will greedily choose which zones to send kits to and ignore some zones with lower variance for too long.  We can circumvent this problem by ensuring that a certain portion of the testing kits, $\rho^{test}$, are distributed proportionally to the population of each zone, and the remaining $1-\rho^{test}$ testing kits are chosen greedily.  This policy falls into the CFA class of policies because it has an embedded optimization problem, and the parameter $\rho^{test}$ must be tuned to find the best proportion on average.  The greedy policy and the proportional policy are given by
\bn
X^{greed}(S_{t}^{cont}|\rho^{test}) =  \min_{\xbf^{test}} (\xbf^{test})^T D \xbf^{test} + d^T \xbf^{test},
\en
\bns
\text{subject to} \hspace{3mm} \mathbf{1}^T \xbf^{test} \leq n_t^{test} - \lfloor \rho^{test}n_t^{test}\rfloor.
\ens
and,
\bns
X^{prop}(S_t^{cont}|\rho^{test}) = \left[\lfloor \rho^{test} n^{test}_t \frac{N_z}{\sum_{z\in\Zcal}N_z} \rfloor\right]_{z\in\Zcal},
\ens
respectively.  Hence, the resulting testing policy becomes,
\bn
X^{test}(S_t^{cont}|\rho^{test}) = X^{greed}(S_{t}^{cont}|\rho^{test}) + X^{prop}(S_{t}^{cont}|\rho^{test}).
\en
The best $\rho^{test}$ in this function class can be tuned through policy evaluation using the simulator defined in section \ref{section:env_model}.  

\section{Results}
The environment agent model from section \ref{section:env_model} was used to construct a simulator in python.  There were two simulators constructed: one for a toy problem with 25 randomly generated zones in a region discussed in section \ref{section:toy}, and one for a simulation of the states in the United States discussed in section \ref{section:USA}.  Considering a vaccine would not be developed until a country is in the midst of a pandemic, the initial state of the infected population is randomly generated from a uniform distribution between $5\%$ and $15\%$ at $t=0$.  The testing and vaccine manufacturing process is assumed to be increasing (on average) to simulate the reality that there will not be a large number of doses at the onset of its distribution into the population.  

\subsection{Toy Problem Simulator}
\label{section:toy}
The toy problem simulator was constructed by randomly generating 25 sample locations, which would be considered nodes of a graph, and then weighting the mobility transition matrix, $\Mhat_{t+1,zz'}$, by the distance between the locations.  Hence, individuals from nearby zones will have a higher probability of interacting with one another.  Each zone has its own transmission rate and the removal rate for the disease remains constant for each zone.  The graph of the simulated region is shown in Figure 1.  The weights of the edges of the graph represent the probability of individuals from each zone interacting with one another.  
\begin{figure}[ht]
\includegraphics[width=0.3\textwidth, center]{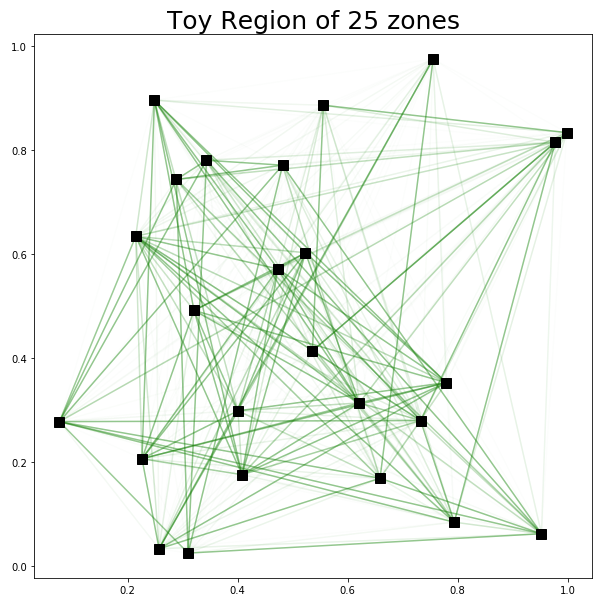}
\caption{The graph of the simulated region with weighted edges proportional to average transmission rates between zones}
\label{fig:cumulative}
\end{figure}

Each of the policies in section \ref{section:policy} were implemented and the results of the cumulative total of infected individuals in each zone can be seen in figure \ref{fig:cumulative}.  The null policy shows the amount of infected individuals if there was no vaccine available with constant transmission rates.  After tuning the parameters of the PFA using a discretized grid search over 50 simulations for each value, the best values were $\theta^{PFA} = (50,10)$. After tuning the risk parameter in the DLA, the best value after 50 simulations for each was $\theta^{risk}= 0.15$.  The active learning policy was tuned through policy evaluation and $\rho^{test} = 0.7$ after 50 simulations for both policies.
\begin{figure}[ht]

\begin{subfigure}{0.5\textwidth}
\includegraphics[width=0.9\linewidth, height=5cm]{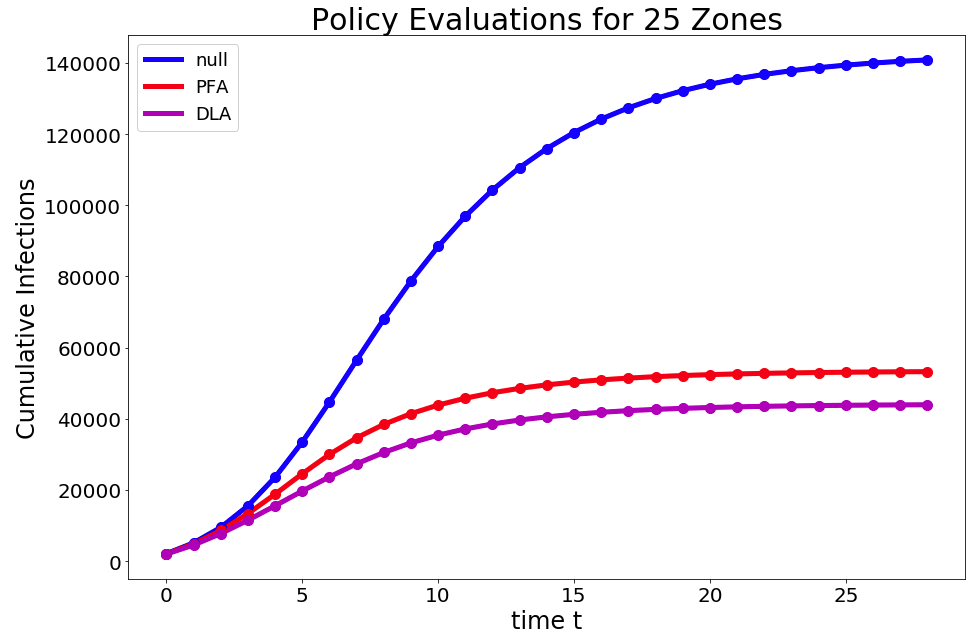} 
\caption{Cumulative number of infected individuals for each policy after 30 weeks}
\label{fig:subim1}
\end{subfigure}
\begin{subfigure}{0.5\textwidth}
\includegraphics[width=0.9\linewidth, height=5cm]{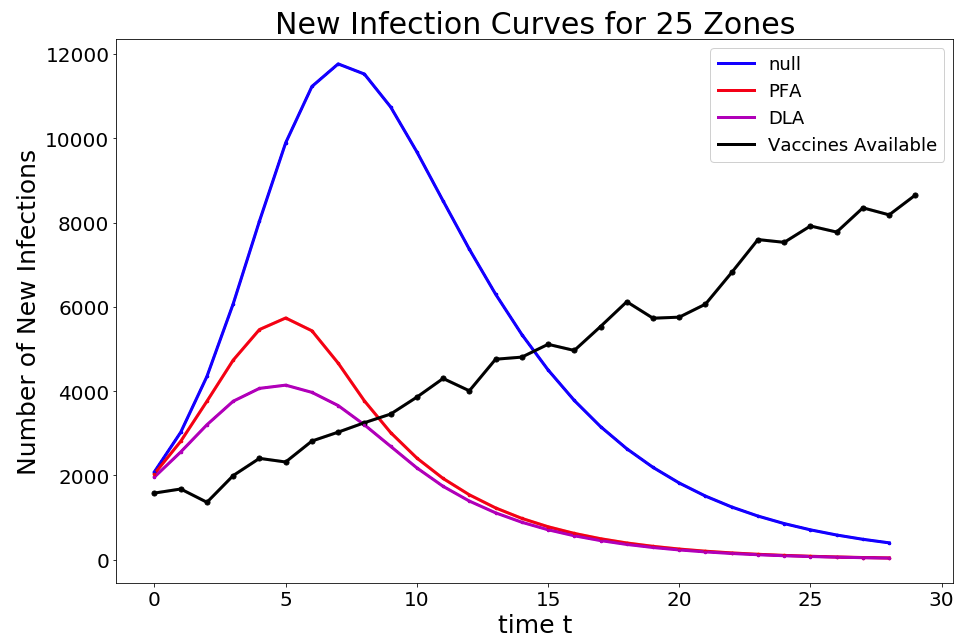}
\caption{New infections each week for 30 weeks accompanied by the weekly number of vaccines available}
\label{fig:subim2}
\end{subfigure}

\caption{Infection totals for each of the policies averaged over 100 sample paths}
\label{fig:image2}
\end{figure}

It is clear that each of the policies offers a significant improvement over the 140,000 people who would be infected if there was no vaccine being distributed in this region.  The tuned PFA offers a $62.2\%$ reduction in cases over the null policy and the DLA offers a $69\%$ reduction in cases over the null policy.

\subsection{USA COVID-19 Simulator}
\label{section:USA}
In the model of the United States, the zones in the country are partitioned at the state level.  The population data for each state was provided by the US Census Bureau.  Each state in the country has a known population, $N_z$.  The probability matrix, $\Mhat_{t+1,zz'}$, for each state was weighted by the distance between each state because neighboring states are more likely to infect each other.  The beta values in each state are created by taking the logarithm of the population density and rescaling it between a fixed range of $(\beta_{min}, \beta_{max})$.  The policies discussed in section \ref{section:policy} were implemented with $|\Zcal| = 51$ zones (including Washington, DC).  The initial infection rate was generated by the record day of daily new infection data in each state to demonstrate that this strategy could work at the peak of a pandemic.
\begin{figure}[ht]
\includegraphics[width=0.5\textwidth, center]{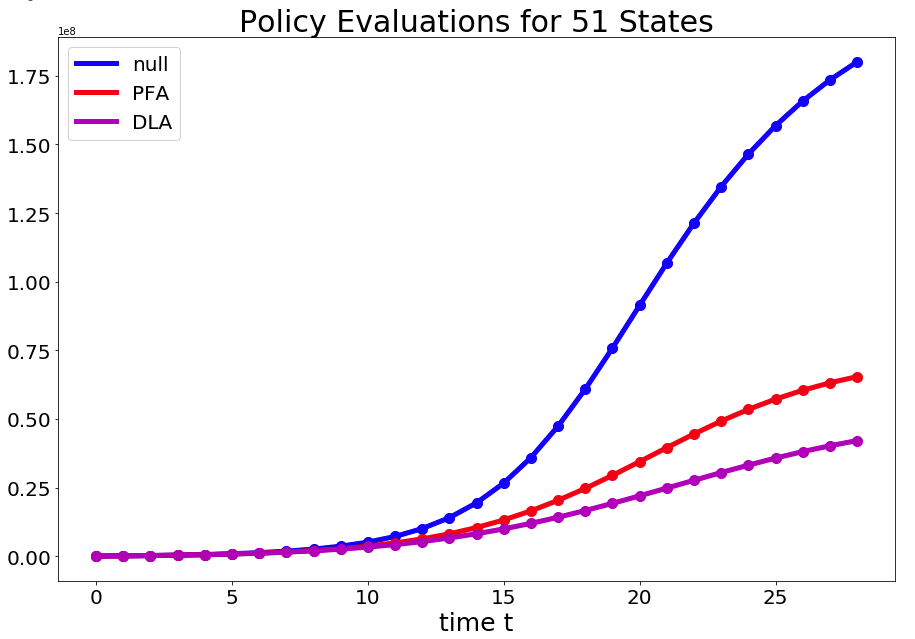}
\caption{The cumulative infections for the 50 states (plus Washington, DC) for both of the policies}
\label{fig:US_cumul}
\end{figure}

In the USA simulator, we assume that the vaccines are being manufactured and the initial vaccine total is about $1\%$ of the population.  As the manufacturing capabilities increase, the total vaccines available for distribution will increase stochastically.  Under the null policy with the assumed $\beta_z$ values, the simulator predicts about 18 million people would be infected with the disease.  The PFA policy improves on the null policy by $45\%$ and the DLA improves on the null policy by $54\%$.  The DLA policy will also prevent about 1.8 million more infections than to the PFA policy over the given time horizon.

\begin{figure}[ht]
\includegraphics[width=0.9\textwidth, center]{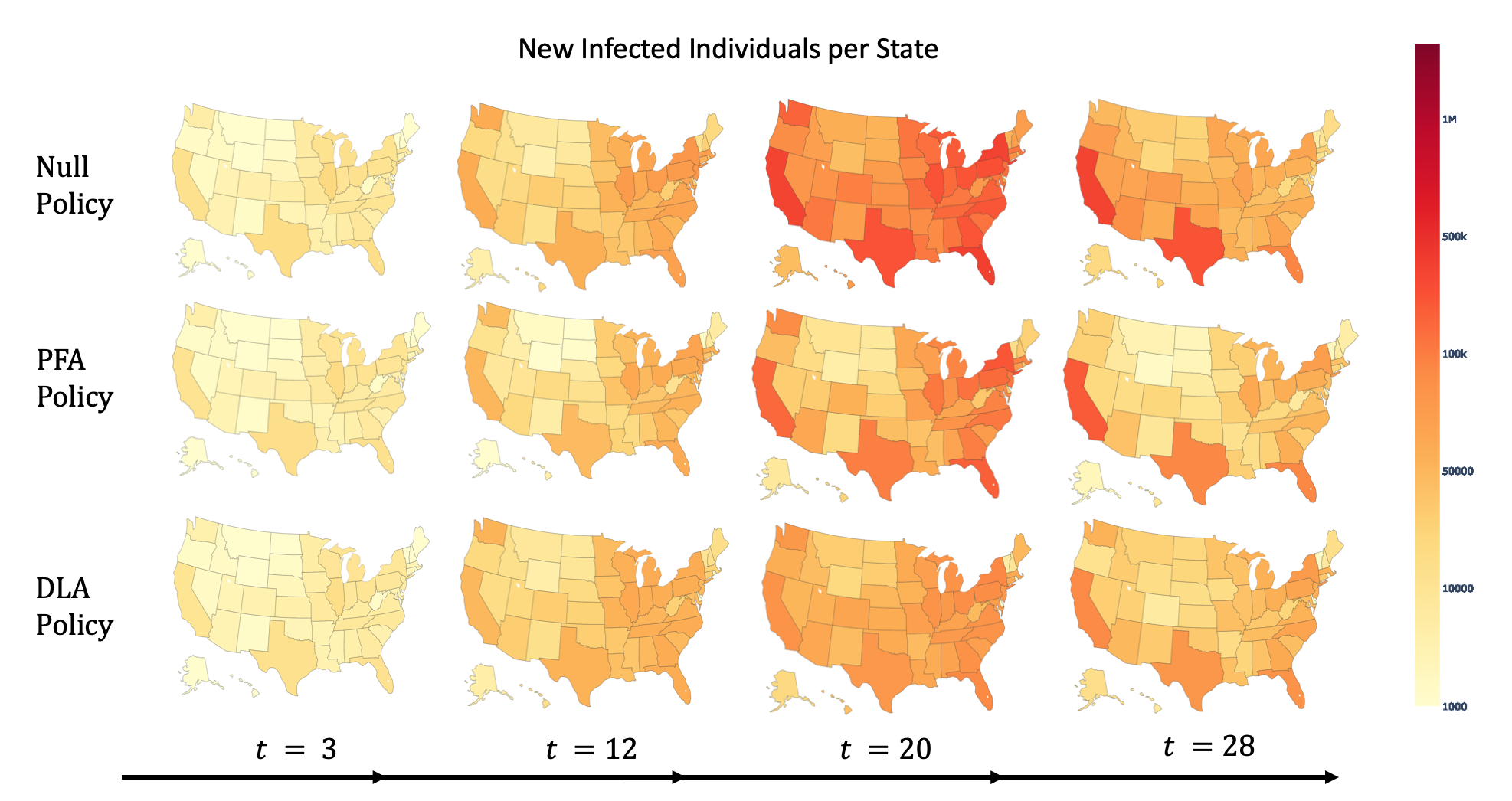}
\caption{New infected individuals per state for three policies: the null policy, PFA, and DLA}
\label{fig:US_graphs}
\end{figure}

When the problem is scaled up to 51 zones and the populations grow to be on the order of millions of people, then the number of infected individuals has the potential to grow significantly larger.  The number of susceptible people determines the rate at which the infected group will grow, so it is important to allocate vaccines strategically in areas with large populations of susceptible individuals.  In the plot in figure \ref{fig:US_graphs}, we show a breakdown of the country as the pandemic spreads with three very different scenarios.  In the first row, there are no vaccines so the infection can spread uncontrollably.  In the second row, the PFA strategy is more likely to spread the vaccine evenly among states if there is a high percentage of susceptible individuals; hence the states with lower populations and transmission rates are targeted with more vaccines.  The third row shows the DLA policy which is much more aggressive towards states with higher population density.  It is obvious that states like New York, California, and Illinois, which have large cities, will spread the virus within their own state at a higher rate than states like Wyoming or Montana. Furthermore, many individuals travel within those high population states, so when the DLA targets those states with vaccines it will have higher order effects.  Hence, the DLA is capable of mitigating those effects.

It is intuitive that early action will always produce the fewest number of infections in the population, which is also why an aggressive DLA has better results.  If we assume that the death rate due to the disease is fixed throughout the time horizon, then it is just a linear function of the number of infected individuals.  Therefore, every infection prevented will be proportional to the number of deaths prevented from the disease.  The risk-adjusted DLA policy prevented 1.8 million more infections than the PFA policy.  Hypothetically, if a $1\%$ death rate is assumed, then the DLA policy would save about 180,000 more lives than the myopic PFA.

\section{Conclusion}
This work presented a formal model for managing vaccines and testing kits during a pandemic using the unified framework.  During a pandemic, the controller would not know the true state of the number of susceptible, infected and removed individuals; hence, the belief modeling in this paper captures the belief states and belief transitions.  The multiagent modeling strategy in this paper is used to distinctly separate the modeling of the environment from the modeling of the controller.  The strategy allows the environment to be considered a blackbox and the controller can only query observations and make decisions which impact it, which is why we focus on a tunable policy which can be optimized to mitigate the risks of an uncertain environment.  Hence, if a more advanced environment model is applied to the problem, then the controller would still be able to robustly be applied.

There were two types of decisions in this paper: the implementation decision and the learning decisions.  This paper compared two implementation policies in the PFA class and the DLA class while using a CFA active learning policy for the testing decisions.  The results demonstrate that the DLA class with a tuned risk adjustment parameter will perform the best in the toy problem and the US states problem.  There are many more complex policies which could be applied to this problem, but we selected two that would illustrate the concepts most clearly.  The PFA policy is intended to illustrate a very basic policy which is more likely to be used in practice than the DLA.  The DLA policy is obviously more complex because it uses lookahead modeling and optimization, while the PFA is an analytical function.  Therefore, we demonstrate that a well designed DLA policy can significantly outperform the basic policies most likely to be used in practice by myopic decision-makers.  

\section{Bibliography}

\bibliography{mybibfile}

\newpage
\section{Appendix}
\paragraph{\bf{Lemma \ref{lem:bias_lemma}}}
\begin{proof}
This proof uses definitions \ref{def: Prob}-\ref{def:cd}.  Since each of the probabilities are impacted by the number tests allocated there is an implicit assumption that $x_{tz}^{test}$ is fixed.  Hence, the probability of getting a test given that the person is positive to be written as,
\bns
\Pbb[E^{test}|E^{pos}] &=& \Pbb[E^{test}|E^{symp},E^{pos}]\Pbb[E^{symp}|E^{pos}] + \Pbb[E^{test}|E^{asymp},E^{pos}]\Pbb[E^{asymp}|E^{pos}] \\
&=& a\Pbb[E^{test}|E^{symp},E^{pos}] + (1-a)\Pbb[E^{test}|E^{asymp},E^{pos}]
\ens
To find those probabilities we can use the assumptions given by the definitions implicit in the environment, hence
\bns
\Pbb[E^{test}|E^{symp}] &=& \Pbb[E^{test}|E^{symp},E^{pos}]\Pbb[E^{pos}|E^{symp}] + \Pbb[E^{test}|E^{symp},E^{neg}]\Pbb[E^{neg}|E^{symp}] \\ 
&=& \Pbb[E^{test}|E^{symp},E^{pos}](\Pbb[E^{pos}|E^{symp}] + \Pbb[E^{neg}|E^{symp}]) \\
&=& \Pbb[E^{test}|E^{symp},E^{pos}] = c, \\
\Pbb[E^{test}|E^{asymp}] &=& \Pbb[E^{test}|E^{asymp},E^{pos}]\Pbb[E^{pos}|E^{asymp}] + \Pbb[E^{test}|E^{asymp},E^{neg}]\Pbb[E^{neg}|E^{asymp}] \\ 
&=& \Pbb[E^{test}|E^{asymp},E^{pos}](\Pbb[E^{pos}|E^{asymp}] + \Pbb[E^{neg}|E^{asymp}]) \\
&=& \Pbb[E^{test}|E^{asymp},E^{pos}] = d.
\ens
Now, those values can be substituted back into the original equation for the probability of getting a test given that the person is positive,
\bns
\Pbb[E^{test}|E^{pos}] &=&  a\Pbb[E^{test}|E^{symp},E^{pos}] + (1-a)\Pbb[E^{test}|E^{asymp},E^{pos}] \\
&=&  ac + (1-a)d.
\ens

The same logic is followed for the probability of getting a test given that the person is negative,
\bns
\Pbb[E^{test}|E^{neg}] &=& \Pbb[E^{test}|E^{symp},E^{neg}]\Pbb[E^{symp}|E^{neg}] + \Pbb[E^{test}|E^{asymp},E^{neg}]\Pbb[E^{asymp}|E^{neg}] \\
&=& b\Pbb[E^{test}|E^{symp},E^{neg}] + (1-b)\Pbb[E^{test}|E^{asymp},E^{neg}] \\
&=& b\Pbb[E^{test}|E^{symp},E^{pos}] + (1-b)\Pbb[E^{test}|E^{asymp},E^{pos}] \\
&=& bc + (1-b)d
\ens
Furthermore, the marginal probability of getting a test is given by,
\bns
\Pbb[E^{test}] &=& \Pbb[E^{test} | E^{pos}]\Pbb[E^{pos}] + \Pbb[E^{test}| E^{neg}]\Pbb[E^{neg}]\\
&=& (ac + (1-a)d) p^{inf}_{tz} + (bc + (1-b)d) (1-p^{inf}_{tz}).
\ens
$p^{test}_{tz}$ is defined as $\Pbb[E^{tpos}|E^{test}, x_{tz}^{test}]$.  Through Bayes Theorem, the probability of truly being positive given that a person goes to get tested is given by,
\bns
\Pbb[E^{pos}|E^{test}] &=& \frac{\Pbb[E^{test}|E^{pos}]\Pbb[E^{pos}]}{\Pbb[E^{test}]}\\
&=& \frac{(ac + (1-a)d) p^{inf}_{tz}}{(ac + (1-a)d) p^{inf}_{tz} + (bc + (1-b)d) (1-p^{inf}_{tz})}.
\ens
These terms can be rearranged to arrive at,
\bns
\Pbb[E^{pos}|E^{test}] =\frac{\left(ac + (1-a)d\right)p^{inf}_{tz}}{(c - d)\left((a-b)p^{inf}_{tz} + b\right) + d}.
\ens

Let $f_p$ and $f_n$ be the probability of false positive and false negative, respectively.  The following equation computes the probability of testing positive given that the person goes to get tested,
\bns
p_{tz}^{test} &=& \Pbb[E^{tpos}|E^{test}, E^{pos}, x_{tz}^{test}]\Pbb[E^{pos}|E^{test}] + \Pbb[E^{tpos}|E^{test}, E^{neg}, x_{tz}^{test}]\Pbb[E^{neg}|E^{test}] \\
&=& (1-f_n) \Pbb[E^{pos}|E^{test}] + f_p (1-\Pbb[E^{pos}|E^{test}]).
\ens

\end{proof}

\paragraph{\bf{Lemma \ref{lem:beta_binomial}}} Let $\alpha$ and $\beta$ be parameters of a beta distribution.  Assume a sample $\Ihat$ from a binomial distribution after $n$ trials and each trail has probability $p$.  The pdf of the prior beta distribution is given by,
\bns
\pi(x|\alpha,\beta) = \frac{1}{B(\alpha,\beta)}x^{\alpha-1} (1-x)^{\beta-1},
\ens
where,
\bns
B(\alpha,\beta) = \frac{\Gamma(\alpha)\Gamma(\beta)}{\Gamma(\alpha+\beta)}.
\ens
The likelihood of observing $\Ihat_{t+1}$ is given by,
\bns
\Pbb\left(\Ihat_{t+1}=k|n ,p\right) = L(p|k) = \left(\begin{matrix} n \\ k \end{matrix}\right) p^{k} (1-p)^{n-k}.
\ens
Therefore, the Bayesian update is given by,
\bns
\Pbb[p|\alpha,\beta] &=& \frac{L(p=x|k) \pi(x|\alpha,\beta)}{\int_{0}^{1} L(p=x|k) \pi(x|\alpha,\beta)dx} = \frac{ \left(\begin{matrix} n \\ k \end{matrix}\right) x^{k} (1-x)^{n-k}\frac{1}{B(\alpha,\beta)}x^{\alpha-1} (1-x)^{\beta-1}}{\int_{0}^{1} \left(\begin{matrix} n \\ k \end{matrix}\right) x^{k} (1-x)^{n-k}\frac{1}{B(\alpha,\beta)}x^{\alpha-1} (1-x)^{\beta-1}dx} \\
&=& \frac{x^{k} (1-x)^{n-k}x^{\alpha-1} (1-x)^{\beta-1}}{\int_{0}^{1} x^{k} (1-x)^{n-k}x^{\alpha-1} (1-x)^{\beta-1}dx} = \frac{x^{k+\alpha-1} (1-x)^{n+\beta-k-1}}{\int_{0}^{1} x^{k+\alpha-1} (1-x)^{n+\beta-k-1}dx}.
\ens
The integral in the denominator is given by,
\bns
\int_{0}^{1} x^{k+\alpha-1} (1-x)^{n+\beta-k-1}dx = B(\alpha+k,n+\beta-k).
\ens
Hence,
\bns
\frac{x^{k+\alpha-1} (1-x)^{n+\beta-k-1}}{B(\alpha+k,n+\beta-k)},
\ens
which is a beta distribution with parameters $\alpha+k$ and $n+\beta-k$.  If $\pi(x|\alpha,\beta)$ was the prior distribution for the value of $p$, then after the observation $\Ihat=k$ the posterior will also have a beta distribution with parameters $\alpha+k$ and $n+\beta-k$.  Therefore, the mean of $p$ given by the posterior distribution is 
\bns
\phat = \frac{\Ihat+\alpha}{n+\alpha+\beta}.
\ens
If X is a random variable with a beta distribution, $X\sim\pi(x|\alpha,\beta)$.  Then its expected value is given by, $\E[X] = \frac{\alpha}{\alpha+\beta}$.  Any samples drawn from this distribution can be thought of as being drawn from a binomial distribution with parameter $p$ being drawn from the prior beta distribution with parameters $\alpha$ and $\beta$.  The terms $\alpha$ and $\beta$ can be thought of as "pseudo - observations" of the successes and failures.  For Lemma \ref{lem:beta_binomial} assume $n=x_{tz}^{test}$, $\alpha_{tz}^{prior} = \alpha$, $\beta_{tz}^{prior} = \beta$, and $\Ihat_{t+1,z}=\Ihat$.

\paragraph{\bf{Lemma \ref{lem:itrans_lem}}} Assume that $(S_{tz}, I_{tz}, R_{tz})$ has a multinomial distribution with parameters $(N_z, \pbar_{tz}^{susc}, \pbar_{tz}^{inf}, \pbar_{tz}^{rec})$.  The subpopulations of each zone are not truly independent; however, when the populations of the zones are large it is a reasonable assumption to make.  Let $Y_{tz} = \min(S_{tz}, x_{tz}^{vac})$.  The percent error introduced into the model by assuming there are no correlations would be given by,
\bns
\epsilon &=& \frac{\left|\E[\min(S_{tz}, x_{tz}^{vac})  I_{tz}|S_{t}^{cont}, \xbf_{t}^{vac}] - \E[\min(S_{tz}, x_{tz}^{vac})|S_{t}^{cont}, \xbf_{t}^{vac}]  \E[I_{tz}|S_{t}^{cont}, \xbf_{t}^{vac}] \right|}{\E[\min(S_{tz}, x_{tz}^{vac})  I_{tz}|S_{t}^{cont}, \xbf_{t}^{vac}]} \\ 
 &=& \frac{\left|Cov[\min(S_{tz}, x_{tz}^{vac}),  I_{tz}|S_{t}^{cont}, \xbf_{t}^{vac}] \right|}{\E[\min(S_{tz}, x_{tz}^{vac})  I_{tz}|S_{t}^{cont}, \xbf_{t}^{vac}]}.
\ens
Let $\epsilon'$ be the percent error by assuming that $S_{tz}$ and $I_{tz}$ are independent random variables given by,
\bns
\epsilon' &=& \frac{\left|Cov[S_{tz},  I_{tz}|S_{t}^{cont}, \xbf_{t}^{vac}] \right|}{\E[S_{tz}I_{tz}|S_{t}^{cont}, \xbf_{t}^{vac}]} = \frac{N_z \pbar_{tz}^{susc} \pbar_{tz}^{inf}}{N_z (N_z-1) \pbar_{tz}^{susc} \pbar_{tz}^{inf}} = \frac{1}{N_z-1}.
\ens
If the errors, $\epsilon'$, are equal to $\frac{1}{N_z-1}$, then when the population is large the percent error will get very small.  Consider the case $S_{tz}<x_{tz}^{vac}$, this implies that $Y_{tz} = S_{tz}$; hence, $\epsilon = \epsilon'$.  The other case is when $S_{tz} \geq x_{tz}^{vac}$, this implies that $Y_{tz} = x_{tz}^{vac}$ which is deterministic; hence, $\epsilon = \E[x_{tz}^{vac} I_{tz}|S_{t}^{cont}, \xbf_{t}^{vac}] - x_{tz}^{vac} \E[I_{tz}|S_{t}^{cont}, \xbf_{t}^{vac}]=0$.  This implies that $\epsilon \leq \epsilon'=\frac{1}{N_z-1}$.  Therefore, the percent errors are less than or equal to $\frac{1}{N_z-1}$ which become very small as $N_z$ gets large when independence is assumed; hence it is a reasonable assumption.  

The belief about the environment agent transition functions for time $t+1$ is given by,
\bns
\Sbf^x_{t} &=& (\Sbf_t - \xbf_{t}^{vac})^+, \\
\Sbf_{t+1} &=& \Sbf^x_{t} - \bbf \odot \Sbf_t^x \odot \Ibf_t,\\
\Ibf_{t+1} &=& (1-\gamma) \Ibf_{t} + \bbf \odot \Sbf_t^x \odot \Ibf_t,\\
\Rbf_{t+1} &=& \Rbf_{t} + \gamma \Ibf_t + min(\Sbf_t, \xbf_t^{vac}).
\ens
If we assume that the size of the population is fairly large, then through the central limit theorem we claim that each of the random variables $S_{tz}$, $I_{tz}$, and $R_{tz}$ can be approximated by independent normal distributions with parameters $(\Sbar_{tz}, \sigma_{tz}^{susc}),(\Ibar_{tz}, \sigma_{tz}^{inf})$ and $(\Rbar_{tz}, \sigma_{tz}^{rec})$, respectively.  Assume $S_{tz} \sim N\left(\Sbar_{tz}, (\sigma_{tz}^{susc})^2\right)$, $I_{tz} \sim N\left(\Ibar_{tz}, (\sigma_{tz}^{inf})^2\right)$, and independence, then
\bns
\E[(S_{tz} - x_{tz}^{vac})^+  I_{tz}|S_{t}^{cont}, \xbf_{t}^{vac}]&=&\Sbar_{tz}^x \Ibar_{tz}.
\ens
Through the identity in \cite{Zhan_IE}, the term in the previous equation is given by,
\bns
\E[(S_{tz} - x_{tz}^{vac})^+| S^{cont}_{t}, x_{tz}^{vac}] &=& \left( (\Sbar_{t,z} -x_{tz}^{vac})\Phi\left(\frac{\Sbar_{t,z} -x_{tz}^{vac}}{\sigma^{susc}_{tz}}\right) + \sigma^{susc}_{tz}\phi\left(\frac{\Sbar_{t,z} -x_{tz}^{vac}}{\sigma^{susc}_{tz}}\right)\right).
\ens
Furthermore, consider the identity $(S_{tz}-x_{tz}^{vac})^+ = S_{tz} - \min(S_{tz}, x_{tz}^{vac})$.  We can use this identity to estimate $\E[\min(S_{tz}, x_{tz}^{vac})| S^{cont}_{t}, x_{tz}^{vac}]$ through the linearity of expectation; hence,
\bns
\E[\min(S_{tz}, x_{tz}^{vac})| S^{cont}_{t}, x_{tz}^{vac}] = \Sbar_{tz} - \Sbar_{tz}^x.
\ens
Finally, through the weak law of large numbers, as $N_z \mapsto \infty$ then $\Sbar_{tz} \mapsto \E[S_{tz}| S^{cont}_{t}, x_{tz}^{vac}]$ because it has finite variance.  The other belief state variables have the same result.  Therefore the estimates of the belief state variables when they are assumed to have independent normal distributions are given by,
\bns
\Sbar_{tz} &=& \E[S_{tz} | S^{cont}_{t}, x_{tz}^{vac}] = N_z \pbar_{tz}^{susc}, \\
\Ibar_{tz} &=& \E[I_{tz} | S^{cont}_{t}, x_{tz}^{vac}] = N_z \pbar_{tz}^{inf},  \\
\Rbar_{tz} &=& \E[R_{tz} | S^{cont}_{t}, x_{tz}^{vac}] = N_z \pbar_{tz}^{rec},\\
\sigma_{tz}^{susc} &=& \sqrt{Var[S_{tz}| S^{cont}_{t}, x_{tz}^{vac}]} = \sqrt{N_z \pbar_{tz}^{susc} ( 1- \pbar_{tz}^{susc})}. \\
\ens
We estimate each expectation by assuming the random variables are normally distributed.  Hence,
\bns
\Sbar_{t+1,z} &=& \E[(S_{tz} - x_{tz}^{vac})^+- \frac{\beta_z}{N_z} (S_{tz} - x_{tz}^{vac})^+  I_{tz}|S_{t}^{cont}, \xbf_{t}^{vac}], \\ 
 &=& \E[(S_{tz} - x_{tz}^{vac})^+|S_{t}^{cont}, \xbf_{t}^{vac}] - \frac{\beta_z}{N_z} \E[(S_{tz} - x_{tz}^{vac})^+  I_{tz}|S_{t}^{cont}, \xbf_{t}^{vac}], \\ 
&=& \left(1 - \beta_z \pbar_{tz}^{inf}\right) \Sbar_{tz}^x.\\
\Ibar_{t+1,z} &=& \E[(1-\gamma) I_{tz} + \frac{\beta_z}{N_z} (S_{tz}-x_{tz}^{vac})^+ I_{tz} |S_{t}^{cont}, \xbf_{t}^{vac}], \\
&=& (1-\gamma)\Ibar_{tz} + \beta_z \pbar_{tz}^{inf} \Sbar_{tz}^x.\\
\Rbar_{t+1,z} &=& \E[R_{tz} + \gamma I_{tz} + \min(S_{tz}, x_{tz}^{vac})|S_{t}^{cont}, \xbf_{t}^{vac}],\\
&=& \E[R_{tz}|S_{t}^{cont}, \xbf_{t}^{vac}] + \E[\gamma I_{tz}|S_{t}^{cont}, \xbf_{t}^{vac}] + \E[\min(S_{tz}, x_{tz}^{vac})|S_{t}^{cont}, \xbf_{t}^{vac}]\\
&=& \Rbar_{tz} + \gamma \Ibar_{tz} +\Sbar_{tz} - \Sbar_{tz}^x.
\ens

\paragraph{\bf{Lemma \ref{lem:learning_pol}}} 
\begin{proof}
The optimization problem for the learning policy is given by,
\bns
\min_{\xbf^{test}} \sum_{z\in\Zcal} Var[\pbar_{t+1,z}^{inf}(\xbf^{test})|S^{cont}_t],
\ens
\bns
\text{subject to} \hspace{3mm} \mathbf{1}^T \xbf^{test} \leq n_t^{test}.
\ens
The objective function for the learning policy is given by equation (\ref{eqn:beta_bin_var}).  The objective function can be rewritten as,
\bns
\sum_{z\in \Zcal} Var[\pbar_{t+1,z}^{inf}|S^{cont}_t] = \sum_{z\in \Zcal}\frac{x^{test}_{tz} \alpha_{tz}^{prior} \beta_{tz}^{prior} ( \alpha_{tz}^{prior} + \beta_{tz}^{prior} + x^{test}_{tz}) }{( \alpha_{tz}^{prior} + \beta_{tz}^{prior})^2 ( \alpha_{tz}^{prior} + \beta_{tz}^{prior} + 1)}.
\ens
Recall that $\alpha_{tz}^{prior} + \beta_{tz}^{prior} = N_z$ for all zones therefore it is a constant.  Which means that,
\bns
\sum_{z\in \Zcal} Var[\pbar_{t+1,z}^{inf}|S^{cont}_t] &=& \sum_{z\in \Zcal}\frac{x^{test}_{tz} \alpha_{tz}^{prior} \beta_{tz}^{prior} ( \alpha_{tz}^{prior} + \beta_{tz}^{prior} + x^{test}_{tz}) }{( \alpha_{tz}^{prior} + \beta_{tz}^{prior})^2 ( \alpha_{tz}^{prior} + \beta_{tz}^{prior} + 1)},\\
&=& \sum_{z\in \Zcal}\frac{x^{test}_{tz} \alpha_{tz}^{prior} \beta_{tz}^{prior} ( N_z + x^{test}_{tz}) }{(N_z)^2 ( N_z + 1)},\\
&=& \mathbf{d}^T \mathbf{x}^{test}_{t} + (\mathbf{x}^{test}_{t})^T D \mathbf{x}^{test}_{t},
\ens
where $\mathbf{d} = \left[\frac{\alpha_{tz}^{prior} \beta_{tz}^{prior}}{N_z ( N_z + 1)}\right]_{z\in\Zcal}$ and $D = diag\left(\left[\frac{\alpha_{tz}^{prior} \beta_{tz}^{prior}}{N_z^2 ( N_z + 1)}\right]_{z\in\Zcal}\right)$.  The constraint does not change.

\end{proof}

\subsection{DLA policy solution}
\label{section:DLAsolver}
\begin{lemma}
Let $\mathbf{\xtilde}_t = \begin{bmatrix} \mathbf{\xtilde}_{tt} \\ \mathbf{\xtilde}_{t,t+1} \end{bmatrix}$ be a decision vector  where the first $|\Zcal|$ entries are the lookahead decision vectors for time $t$ and the last $|\Zcal|$ entries are the lookahead decision vectors for time $t+1$.  The solution to the two-step lookahead problem in Equation (\ref{eqn:2step_obj}) becomes a quadratic program if $\mathbf{1}^T\mathbf{\Stilde}^{\theta}_{tt} > n^{vac}_t$, otherwise the solution is $\xbf_t^{DLA} = \mathbf{\Stilde}^{\theta}_{tt}$.  The quadratic program for the non-trivial solution is given by,
\bn
\argmax_{\mathbf{\xtilde}_{t}} \hspace{2mm} \mathbf{\xtilde}_t^T Q \mathbf{\xtilde}_{t} + \mathbf{q}^T \mathbf{\xtilde}_t
\en
\bns
\text{subject to} \hspace{3mm} K \xtilde_t \leq \mathbf{h}
\ens
where,
\bns
Q &=& \begin{bmatrix} diag([\bbf \odot \ubf \odot \wbf) & diag([\frac{1}{2} \bbf \odot \ubf]) \\ diag([\frac{1}{2} \bbf \odot \ubf]) & \mathbf{0} \end{bmatrix}, \\ 
\mathbf{q} &=& \begin{bmatrix} (2-\gamma)\ubf + \bbf \odot (\ubf \odot \zbf + \wbf \odot \vbf)  \\ -\bbf \odot \vbf \end{bmatrix}, \\
K &=& \begin{bmatrix} I_{|\Zcal|} & \mathbf{0} \\ -\wbf & I_{|\Zcal|} \end{bmatrix}, \\ 
\mathbf{h} &=& \begin{bmatrix}\mathbf{\Stilde}^{\theta}_{tt}  \\ \zbf \end{bmatrix}, \\
\ens
where each term is $\bbf = \left[\frac{\beta_z}{N_z}\right]_{z\in\Zcal}$, $\ubf = \left[-\frac{\beta_z}{N_z}\Itilde_{ttz}\right]_{z\in\Zcal}$, $\wbf = \left[\left(\frac{\beta_z}{N_z}\Itilde_{ttz}-1\right)\right]_{z\in\Zcal}$, $\vbf = \left[(1-\gamma)\Itilde_{ttz} + \frac{\beta_z}{N_z} \Stilde^{\theta}_{ttz} \Itilde_{ttz}\right]_{z\in\Zcal}$, 
$\zbf = \left[\Stilde^{\theta}_{ttz} - \frac{\beta_z}{N_z} \Stilde^{\theta}_{ttz} \Itilde_{ttz}\right]_{z\in\Zcal}$, and $I_{|\Zcal|}$ is the identity matrix.
\begin{proof}
Let $\Stilde_{tt}^{cont} = (\Stilde^{\theta}_{ttz}, \Itilde_{ttz}, \Rtilde^{\theta}_{ttz})$ be the lookahead state variable for time $t$ at time $t$.  The lookahead state variable is a point estimate of the distribution represented by the belief state at a tunable quantile.  Equations (\ref{eqn:bstate_trans_sx_base} - \ref{eqn:bstate_trans_r_base}) describe how each of the lookahead state variables will transition.  The objective function in equation (\ref{eqn:2step_obj}) becomes
\bns
 \Ctilde(\Stilde_{tt}^{cont},\mathbf{\xtilde}_{tt})  + \Ctilde(\Stilde_{t,t+1}^{cont}, \mathbf{\xtilde}_{t,t+1}) = \Itilde_{t,t+1,z} + \Itilde_{t,t+2,z},
\ens
\bns
&=& \sum_{z\in\Zcal}\Itilde_{t,t+1,z} + (1-\gamma) \sum_{z\in\Zcal} \Itilde_{t,t+1,z} + \sum_{z\in\Zcal} \frac{\beta_z}{N_z} \Itilde_{t,t+1,z} (\Stilde_{t,t+1,z} - \xtilde_{t,t+1,z})^+\\
&=&  (2-\gamma) \sum_{z\in\Zcal} \Itilde_{t,t+1,z} + \sum_{z\in\Zcal} \frac{\beta_z}{N_z} \Itilde_{t,t+1,z} (\Stilde_{t,t+1,z} - \xtilde_{t,t+1,z})^+.
\ens
We know from equation (\ref{eqn:bstate_trans_i_base}) that the transition equation for $\Itilde_{t,t+1,z}$ is given by,
\bn
\label{eqn:I1LA}
\Itilde_{t,t+1,z} = (1-\gamma) \Itilde_{ttz} + \frac{\beta_z}{N_z} \Itilde_{ttz} (\Stilde^{\theta}_{ttz} - \xtilde_{ttz})^+.
\en
Since these the lookahead state variables are deterministic point estimates, then including the constraint $\Stilde_{ttz} > \xtilde_{ttz}$ will not change the results of the optimization problem in equation (\ref{eqn:2step_obj}).  The constraint implies $(\Stilde_{ttz} - \xtilde_{ttz})^+ = \Stilde_{ttz} - \xtilde_{ttz}$.  Assume the following vectors are defined by, 
\bns
\bbf &=& \left[\frac{\beta_z}{N_z}\right]_{z\in\Zcal},\\
\ubf &=& \left[-\frac{\beta_z}{N_z}\Itilde_{ttz}\right]_{z\in\Zcal},\\ \wbf &=& \left[\left(\frac{\beta_z}{N_z}\Itilde_{ttz}-1\right)\right]_{z\in\Zcal},\\ 
\vbf &=& \left[(1-\gamma)\Itilde_{ttz} + \frac{\beta_z}{N_z} \Stilde^{\theta}_{ttz} \Itilde_{ttz}\right]_{z\in\Zcal},\\ 
\zbf &=& \left[\Stilde^{\theta}_{ttz} - \frac{\beta_z}{N_z} \Stilde^{\theta}_{ttz} \Itilde_{ttz}\right]_{z\in\Zcal}.
\ens
The sum of the terms in equation (\ref{eqn:I1LA}) can be rewritten as,
\bns
\sum_{z\in\Zcal} \Itilde_{t,t+1,z} = \mathbf{1}^T\vbf + \ubf^T \mathbf{\xtilde}_{tt}.
\ens
The objective function can then be rewritten as,
\bns
&=&  (2-\gamma) (\mathbf{1}^T\vbf + \ubf^T \mathbf{\xtilde}_{tt}) + (\bbf \odot (\vbf + \ubf \odot \mathbf{\xtilde}_{tt}))^T(\zbf + \wbf \odot \mathbf{\xtilde}_{tt}) - (\bbf \odot (\vbf + \ubf \odot \mathbf{\xtilde}_{tt}))^T(\mathbf{\xtilde}_{t,t+1}),
\ens
where we also assume the constrain $\zbf + \wbf \odot \mathbf{\xtilde}_{tt} > \mathbf{\xtilde}_{t,t+1}$.  The terms that do not depend on $\mathbf{\xtilde}_{t,t+1}$ or $\mathbf{\xtilde}_{tt}$ can be removed from the objective function; hence,
\bns
&=&  (2-\gamma) (\ubf^T \mathbf{\xtilde}_{tt}) + (\bbf \odot (\vbf + \ubf \odot \mathbf{\xtilde}_{tt}))^T(\zbf + \wbf \odot \mathbf{\xtilde}_{tt}) - (\bbf \odot (\vbf + \ubf \odot \mathbf{\xtilde}_{tt}))^T(\mathbf{\xtilde}_{t,t+1}),\\
&=&  (2-\gamma) (\ubf^T \mathbf{\xtilde}_{tt}) + (\bbf \odot (\zbf \odot \ubf + \vbf \odot \wbf))^T \mathbf{\xtilde}_{tt} +  \mathbf{\xtilde}_{tt}^T diag(\bbf \odot \ubf \odot \wbf)  \mathbf{\xtilde}_{tt}- (\bbf \odot (\vbf + \ubf \odot \mathbf{\xtilde}_{tt}))^T(\mathbf{\xtilde}_{t,t+1}).
\ens
Let $\mathbf{\xtilde}_t = \begin{bmatrix} \mathbf{\xtilde}_{tt} \\ \mathbf{\xtilde}_{t,t+1} \end{bmatrix}$ be a decision vector  where the first $|\Zcal|$ entries are the lookahead decision vectors for time $t$ and the last $|\Zcal|$ entries are the lookahead decision vectors for time $t+1$.  Then the objective function rearranges to the following equation quadratic function,
\bns
\argmax_{\mathbf{\xtilde}_{t}} \hspace{2mm} \mathbf{\xtilde}_t^T Q \mathbf{\xtilde}_{t} + \mathbf{q}^T \mathbf{\xtilde}_t
\ens
\bns
\text{subject to} \hspace{3mm} K \mathbf{\xtilde_t} \leq \mathbf{h}
\ens
where,
\bns
Q &=& \begin{bmatrix} diag([\bbf \odot \ubf \odot \wbf) & diag([\frac{1}{2} \bbf \odot \ubf]) \\ diag([\frac{1}{2} \bbf \odot \ubf]) & \mathbf{0} \end{bmatrix}, \\ 
\mathbf{q} &=& \begin{bmatrix} (2-\gamma)\ubf + \bbf \odot (\ubf \odot \zbf + \wbf \odot \vbf)  \\ -\bbf \odot \vbf \end{bmatrix}, \\
K &=& \begin{bmatrix} I_{|\Zcal|} & \mathbf{0} \\ -\wbf & I_{|\Zcal|} \end{bmatrix}, \\ 
\mathbf{h} &=& \begin{bmatrix}\mathbf{\Stilde}^{\theta}_{tt}  \\ \zbf \end{bmatrix}. \\
\ens

\end{proof}
\end{lemma}

\end{document}